\pdfoutput=1

\documentclass[11pt]{article}

\usepackage[final]{acl}
\usepackage{times}
\usepackage{latexsym}
\usepackage[T1]{fontenc}
\usepackage[utf8]{inputenc}
\usepackage{microtype}
\usepackage{inconsolata}
\usepackage{graphicx}
\usepackage{booktabs}
\usepackage{subcaption}
\DeclareUnicodeCharacter{221E}{\ensuremath{\infty}}
\usepackage{listings}
\usepackage{xcolor}
\usepackage{tabularx}
\usepackage{enumitem}
\usepackage{caption}
\usepackage{longtable}
\usepackage{amsmath} 

\definecolor{keyword}{RGB}{0,0,150}
\definecolor{string}{RGB}{150,0,0}
\definecolor{comment}{RGB}{0,150,0}
\definecolor{background}{RGB}{240,240,240}

\title{Hierarchical Memory Organization for Wikipedia Generation
}

\author{
 \textbf{Eugene J. Yu\textsuperscript{1,2}}\quad
 \textbf{Dawei Zhu\textsuperscript{1,2}}\quad
 \textbf{Yifan Song\textsuperscript{1,2}}\quad
 \textbf{Xiangyu Wong\textsuperscript{1,2}}\quad
 \textbf{Jiebin Zhang\textsuperscript{1,2}}
 \\
 \textbf{Wenxuan Shi\textsuperscript{3}}\quad
 \textbf{Xiaoguang Li\textsuperscript{3}}\quad
 \textbf{Qun Liu\textsuperscript{3}}\quad
 \textbf{Sujian Li\thanks{Corresponding author.}\textsuperscript{1,2}}
\\
 \textsuperscript{1} 
 School of Computer Science, Peking University \\
\textsuperscript{2} National Key Laboratory for Multimedia Information Processing, Peking University \\
 \textsuperscript{3} Huawei Noah's Ark Lab \\
 \{ejyu, lisujian\}@pku.edu.cn \\
}

\begin{document}
\maketitle

\begin{abstract}
Generating Wikipedia articles autonomously is a challenging task requiring the integration of accurate, comprehensive, and well-structured information from diverse sources. 
This paper introduces the Memory Organization-based Generation (\textbf{MOG}) framework, a novel approach to address these challenges by leveraging a hierarchical memory architecture. MOG extracts fine-grained memory units from web documents, recursively organizes them into a Wikipedia-style hierarchical structure, and uses this structure to guide the generation process. This ensures alignment between memory and the article outline, improving both informativeness and verifiability while minimizing hallucinations. Additionally, a citation module is implemented to enhance traceability by linking every generated sentence to specific memory units. 
Evaluations on our newly created WikiStart dataset demonstrate that MOG outperforms baseline methods in producing informative and reliable articles, making it particularly robust in real-world scenarios.\footnote{The code and resources used in this work are publicly available at: \url{https://github.com/eugeneyujunhao/mog}}
\end{abstract}
\section{Introduction}

Large language models (LLMs)~\citep{brown2020language, achiam2023gpt} show remarkable abilities in understanding and generating human language. The integration of Retrieval-Augmented Generation (RAG) enhances these models by enabling them to access and use real-time information from an external memory. This feature is especially useful for tasks requiring data not encountered during training, such as automating Wikipedia article generation. In Wikipedia generation, the memory consists of web documents relevant to a specific topic. Unlike retrieval tasks like question answering~\citep{karpukhin2020dense}, which focus on identifying precise answers, Wikipedia generation requires organizing information into a coherent and accessible format. This task is similar to other organizational tasks that lack a single optimal solution. For instance, \citet{sanderson1999deriving} organized a collection of documents into a concept hierarchy using salient words and phrases. Similarly, in Wikipedia generation, a hierarchical outline serves a similar purpose, providing a framework for organizing information effectively.

\begin{figure}[t]
    \centering
    \includegraphics[width = 0.85\columnwidth]{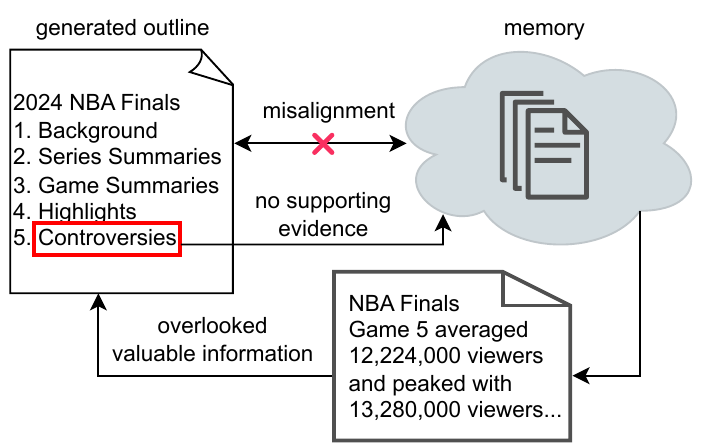}
    \caption{Illustration of misalignment between generated outlines and memory. The generated outline includes a "Controversies" section, despite the absence of related documents in memory. Conversely, a document on media coverage is overlooked by the outline, leading to underutilization of the available memory.}
    \label{fig:problem}
\end{figure}

Due to the extensive context of documents in memory~\citep{qian2023webbrain, gao2024evaluating}, existing methods~\citep{shao-etal-2024-assisting,fan-gardent-2022-generating} do not employ memory directly for outline generation. Instead, they utilize indirect techniques such as inferring context from previously generated sections~\citep{fan-gardent-2022-generating} or using conversation history~\citep{shao-etal-2024-assisting} to construct the outline. This leads to a misalignment between the memory and the outline. As shown in Figure~\ref{fig:problem}, such misalignment can cause sections lacking sufficient supporting evidence, increasing the risk of hallucinations. Additionally, valuable information within the memory may be overlooked, compromising the completeness of the generated article.

\begin{figure}[t]
    \centering
    \includegraphics[width=0.85\columnwidth]{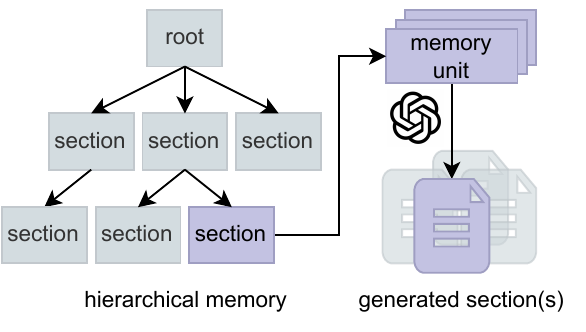}
    \caption{Illustration of the MOG generation process: The memory, equivalent to an outline hierarchy, is composed of section nodes, each associated with non-overlapping memory units.} 
    \label{fig:article_generation}
\end{figure}

To address these challenges, we introduce \textbf{M}emory \textbf{O}rganization-based \textbf{G}eneration (\textbf{MOG}), a memory architecture specifically designed for generating Wikipedia articles. Our approach incorporates fine-grained memory units, each representing a factoid extracted from web documents relevant to the topic. We define basic memory operations (e.g., \textbf{[cluster]}, \textbf{[summarize]}) to utilize these units across multiple levels of information depth. Building on these operations, our approach is divided into three modules: memory construction, memory organization, and article generation.

During the memory construction stage, we enhance information diversity by using a subtopic explorer to retrieve a broader range of documents, inspired by RichRAG~\citep{wang2024richrag}. To maintain relevance, we filter and compile these documents into topic-centric memory units. We then propose a novel algorithm to organize these memory units into a hierarchical structure reflecting a Wikipedia article's layout by recursively clustering and summarizing them. This tree structure dynamically adjusts section depth based on information volume, optimizing memory use and minimizing redundancy. Illustrated in Figure~\ref{fig:article_generation}, this structure serves as a blueprint for article generation, aligning each section with its supporting memory units to reduce hallucination risks. Additionally, a citation module enhances verifiability by linking each sentence to its source memory units. Together, these modules ensure the generated article is accurate, comprehensive, and traceable.

Moreover, many existing evaluation datasets \cite{qian2023webbrain, shao-etal-2024-assisting} focus on high-resource settings with well-developed articles, often overlooking the challenges of real-world scenarios. To address this gap, we introduce WikiStart, a new evaluation dataset derived from Wikipedia articles flagged for expansion. This dataset is designed for low-resource settings, where articles are underdeveloped and require the integration of information scattered across the internet. Additionally, we assess MOG's effectiveness in high-resource settings using the FreshWiki dataset~\citep{shao-etal-2024-assisting}.

Our contributions can be summarized as follows:

\begin{enumerate}
    \item We introduce the MOG paradigm, which transforms relevant web documents into fine-grained memory units and organizes them into a hierarchical outline mirroring the section-based structure of Wikipedia articles.
    \item We propose a novel recursive algorithm to construct this hierarchical outline, ensuring each section is directly supported by evidence, thereby improving accuracy and reducing hallucinations.
    \item We conduct extensive experiments on the FreshWiki dataset and our newly constructed WikiStart dataset, demonstrating that MOG achieves state-of-the-art performance in both informativeness and verifiability.
\end{enumerate}


\begin{table*}[htbp!]
\centering
\begin{tabular}{|l|l|l|}
\hline
\textbf{Operation}      & \textbf{Input \& Output}           & \textbf{Description}                                    \\ \hline
\textbf{[save]}         & $m, l \rightarrow \emptyset$       & Tags a memory unit $m$ with label $l$ and saves it.     \\ \hline
\textbf{[recall]}       & $l \rightarrow \{m\}$             & Retrieves all memory units labeled with $l$.           \\ \hline
\textbf{[extract]}      & $t, d \rightarrow \{m\}$          & Extracts memory units related to topic $t$ from document $d$. \\ \hline
\textbf{[cluster]}      & $\{m\} \rightarrow \{C\}$         & Groups memory units into clusters $\{C\}$.                 \\ \hline
\textbf{[summarize]}    & $C \rightarrow s$                 & Summarizes a memory cluster $C$ into concise text $s$.  \\ \hline
\end{tabular}
\caption{Basic memory operations. $t$ represents a topic, $d$ represents relevant web documents, $m$ is a memory unit, $C=\{m\}$ is a memory cluster, $s$ is a summary, and $l$ is a label (section heading). The symbol $\emptyset$ denotes that the operation produces no output (i.e., it performs a side effect only).}
\label{tab:memory_operations}
\end{table*}

\section{Related Work}
\subsection{External Memory of LLMs}
LLMs possess extensive world knowledge embedded during training. This knowledge enables them to perform few-shot or zero-shot reasoning~\citep{brown2020language, kojima2022large} and tackle complex problems that may involve information beyond their training data. New and task-specific information is temporarily incorporated into the LLM's context window, known as the working context. However, for many complex tasks, the required information may exceed the model's context window, necessitating the use of external memory. In this paper, we define memory as the external memory.

In some applications, memory takes the form of a list of memory objects to simulate an agent’s past experiences, as seen in \citet{park2023generative}. Other approaches, such as \citet{hu2023chatdb}, leverage symbolic memory organized through SQL statements, while \citet{modarressi2023ret} focus on extracting information triplets as memory units for question answering tasks. These methods typically treat memory as a collection of potentially relevant information, from which the most pertinent units are retrieved when needed. In contrast, the role of memory differs in tasks like Wikipedia article generation. In this scenario, each memory unit contains useful content that directly contributes to the final output, which necessitates not just retrieval but also thoughtful organization. In our work, we emphasize structuring this memory into a hierarchical format that reflects the layout of Wikipedia.

\subsection{Memory Construction in Wikipedia Generation}

Generating Wikipedia content requires synthesizing complex information from various internet sources into a structured and coherent format \citep{sauper-barzilay-2009-automatically}. This process typically involves efficiently collecting topic-specific information through search engine queries that fetch relevant web documents. For example, \citet{sauper-barzilay-2009-automatically} and \citet{fan-gardent-2022-generating} create queries by pairing the topic with section headings. Meanwhile, \citet{shao-etal-2024-assisting} devise multi-perspective question scenarios to generate diverse queries. The documents retrieved through these queries serve as a memory foundation, offering supporting evidence for article creation. MOG uses a subtopic explorer to gather documents, then takes one step further to extract fine-grained, relevant factoids to construct memory units instead of using entire documents.

\subsection{Structured Organization in Wikipedia Article Generation}

Creating an effective outline is crucial for organizing gathered information into logical subdivisions and requires meticulous planning. Earlier research \citep{sauper-barzilay-2009-automatically, banerjee-mitra-2015-wikikreator} relied on predefined outline templates for specific categories of topics. In contrast, more recent approaches leverage the advanced planning capabilities of LLMs \citep{liu2018generating, fan-gardent-2022-generating, shao-etal-2024-assisting} to generate outlines for a wider range of topics. These outlines guide the LLMs in crafting articles typically by generating each section individually and subsequently concatenating the sections into a comprehensive article \citep{fan-gardent-2022-generating, shao-etal-2024-assisting}. During the generation of each section, relevant information is retrieved from the unorganized memory through Retrieval-Augmented Generation (RAG) \citep{guu2020retrieval, gao2023retrieval, zhang2024bench} to provide necessary context. To further optimize information utilization, MOG directly leverages memory to develop outlines in a top-down manner. MOG preallocates memory units to non-overlapping hierarchical sections, ensuring each is supported by valid sources, thereby minimizing the risk of hallucination and reducing redundancy.

\begin{figure*}[t]
    \centering
    \begin{subfigure}[b]{0.43\textwidth}
        \centering
        \includegraphics[width=\textwidth]{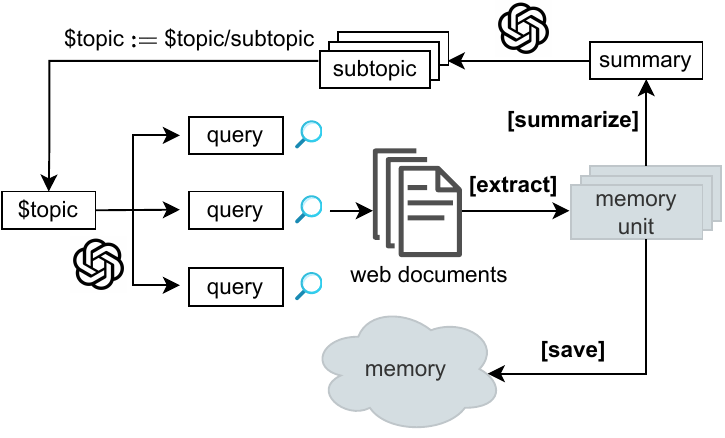}
        \caption{memory construction}
        \label{fig:method_a}
    \end{subfigure}
    \hfill
    \begin{subfigure}[b]{0.53\textwidth}
        \centering
        \includegraphics[width=\textwidth]{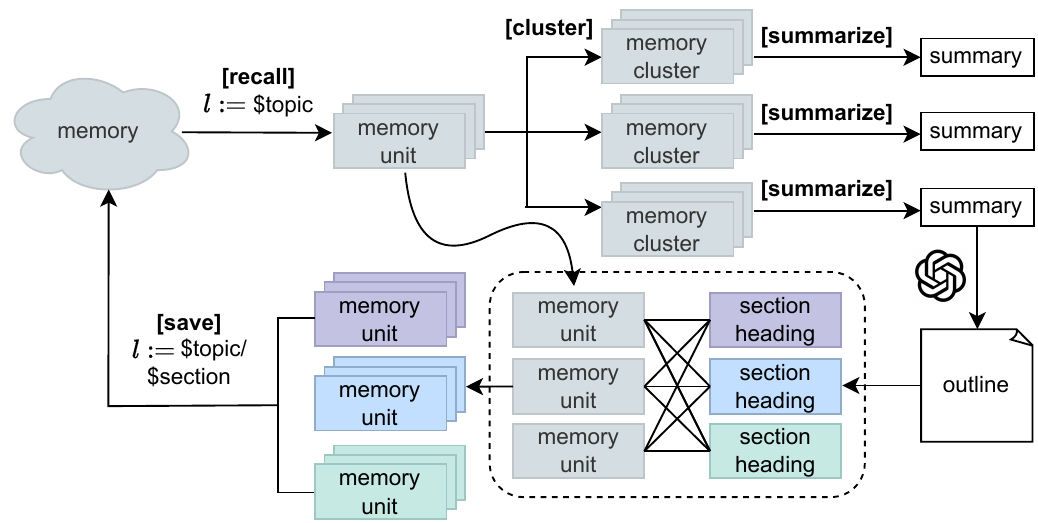}
        \caption{memory organization}
        \label{fig:method_b}
    \end{subfigure}
    \caption{Overview of the method for constructing a hierarchical memory. (a) Memory units are collected, and the article topic is expanded into subtopics during the memory construction stage. (b) These memory units are then recursively organized into a hierarchical structure. This process transforms the memory into an outline hierarchy to support article generation as illustrated in Figure~\ref{fig:article_generation}.} 
    \label{fig:method}
\end{figure*}

\section{Method}
\label{sec:method}
To align with the structural guidelines detailed in the Wikipedia Manual of Style\footnote{\url{https://en.wikipedia.org/wiki/Wikipedia:Manual_of_Style/Layout}}, we have developed a specialized memory architecture. At its core, a memory unit $m$ represents a self-contained factoid expressed concisely in natural language~\citep{chen2023dense}. A memory set, $M=\{m\}$, thus collects all factoids relevant to a given topic. Unlike methods that use document chunks as the basic memory unit—often combining unrelated details and introducing ambiguity through coreferences—our approach focuses on isolating precise, fact-level information to enhance accuracy and reduce redundancy.

\subsection{Memory Operations}
\label{sec:memory_ops}
We propose five operations, described in Table~\ref{tab:memory_operations}, to efficiently manage these memory units. The \textbf{[save]} operation annotates each memory unit with a corresponding Wikipedia section heading and stores both the text and embedding of the memory unit in a vector database.\footnote{We use ChromaDB for this purpose.} The \textbf{[recall]} operation retrieves the memory units based on these headings, allowing us to \textbf{[cluster]} them into thematically relevant groups using the K-Means algorithm based on their embeddings. Additionally, the \textbf{[extract]} operation selects pertinent factoids from the documents, and the \textbf{[summarize]} operation condenses each memory cluster into a concise summary.\footnote{Both \textbf{[extract]} and \textbf{[summarize]} are performed by prompting LLMs.} Together, these operations support the execution of MOG's three main modules: memory construction, memory organization, and article generation.

\subsection{Dynamic Memory Construction}
\label{sec:memory_construction}
Two main issues arise in the memory construction process: determining \textit{what to search} for and understanding \textit{what information is available}. The first involves developing a strategy for curating queries, while the second requires adapting to search results, whether by delving deeper or abandoning a particular line of inquiry. To enhance information collection, we implement a subtopic explorer inspired by RichRAG's sub-aspect explorer~\citep{wang2024richrag}, facilitating the gathering of information on multiple aspects of a topic. 

As shown in Figure~\ref{fig:method}(a), the topic is transformed into search queries and relevant documents are retrieved through a search engine. We then \textbf{[extract]} memory units $\{m\}$ from each document $d$, which are then \textbf{[saved]} into memory $M$. After the initial collection, the \textbf{[summarize]} operation condenses the memory units, and the resulting summary $s$ is used to develop subtopics. The explorer then investigates these subtopics using the same sequence of operations. This process continues until a maximum depth is reached or there is insufficient data.

\subsection{Hierarchical Memory Organization}
The memory consists of a large number of memory units, making it impractical to directly incorporate all of them within the context window of a LLM to generate a comprehensive outline. To address this, we propose a recursive organization of memory from coarse-grained to fine-grained levels. 

As illustrated in Figure~\ref{fig:method}(b), the initial step involves \textbf{[clustering]} memory units $\{m_l\}$ labeled as section $l$ into groups $\{C\}$ based on their semantic similarity. Each cluster is then \textbf{[summarized]} to create a concise representation of its contents. These summaries $\{s\}$ serve as input for the LLM to generate the subsection headings $\{l'\}$, effectively structuring the outline. The individual memory units are subsequently assigned to their corresponding sections by evaluating the semantic similarity of their embeddings to the section headings. 
\begin{equation}
l^* = \arg\max_{l'} \text{sim}(m, l')
\label{eq:assign}
\end{equation}
Specifically, each memory unit $m$ is allocated to the section $l^*$ in which it shares the highest semantic similarity, as determined by Equation~\ref{eq:assign}.

For sections that require more detailed information, the same clustering and summarization process is recursively applied, allowing the outline to expand into subsections with greater specificity. When generating first-level headings, clustering is performed across the entire memory to provide a broad overview, resulting in coarse-grained summaries. As the outline progresses to deeper levels, such as second-level subsections, clustering focuses on the memory units within each specific section, producing finer-grained summaries and deeper insights. This strategy ensures that high-level sections capture the overall structure of the information, while deeper sections delve into detailed aspects. By allowing for flexible adjustments in section depth based on the available information, this approach enhances memory efficiency without neglecting any crucial data.

\subsection{Article Generation}


Following \citet{fan-gardent-2022-generating, shao-etal-2024-assisting}, we adopt a section-by-section generation approach. As illustrated in Figure~\ref{fig:article_generation}, the process begins by extracting the hierarchical outline from the organized memory. For each section, we prompt the LLM to generate coherent content based on the corresponding heading $l$ and its associated memory units $\{m_l\}$. These generated sections are then combined hierarchically according to the outline to form a cohesive article. This is followed by a final polishing step that ensures overall coherence and consistency across the entire document.

Furthermore, we incorporate a post-hoc citation module to address concerns regarding the trustworthiness of generated content. Benefiting from our generation process which is based on fine-grained memory units, we are able to provide citations supported by precise and granular information~\citep{zhang2024longcite}. We utilize NLTK~\citep{bird2006nltk} to segment the generated content into individual sentences, with each sentence paired with the most relevant memory units using LLM. By retaining source documents during memory construction, this citation mechanism enables precise traceability to the original sources. Consequently, the generated content can be easily verified, allowing straightforward access to the original documents for further validation.
\section{Evaluation}
\subsection{Evaluation Dataset}
We introduce WikiStart, a new evaluation dataset derived from Wikipedia articles flagged for expansion. WikiStart poses distinctive challenges because the required information is often dispersed across various web sources that have not heavily cited the target Wikipedia article. This setting reduces the risk of data leakage and more accurately reflects real-world conditions, as nearly half of Wikipedia entries are stubs in need of substantial expansion.\footnote{\url{https://en.wikipedia.org/wiki/Wikipedia:Stub}} To ensure that the selected topics reflect this condition—where the articles are not already well-established—we retain only relatively short entries that are likely to demand more extensive information gathering. This filtering process results in 1,246 Wikipedia topics (see Appendix~\ref{appendix:dataset} for details). To further demonstrate the versatility of our approach, we also evaluate on FreshWiki~\citep{shao-etal-2024-assisting}, a high-resource dataset of 100 topics with well-developed reference articles. Due to budget constraints and to align with the FreshWiki setup, we sampled 100 topics from WikiStart as our evaluation set.

\begin{table*}[ht] \centering

    \makebox[\textwidth]{\small (a) Results on FreshWiki.} 
    \vspace{1em}
    \resizebox{\textwidth}{!}{
        \Large
        \begin{tabular}{l||*{4}{c}||*{3}{c}||*{4}{c}}
        \toprule
        & \multicolumn{4}{c||}{\textbf{Informativeness}} & \multicolumn{3}{c||}{\textbf{Verifiability}} & \multicolumn{4}{c}{\textbf{Comparison with Human-written Articles}}  \\
        & \begin{tabular}[c]{@{}c@{}}Section \\ Count\end{tabular} & \begin{tabular}[c]{@{}c@{}}Word \\ Count\end{tabular} & \begin{tabular}[c]{@{}c@{}}Entity \\ Count\end{tabular} & \begin{tabular}[c]{@{}c@{}}Numerical \\ Count\end{tabular} & \begin{tabular}[c]{@{}c@{}}Citation \\ Recall\end{tabular} & \begin{tabular}[c]{@{}c@{}}Citation \\ Precision\end{tabular} & \begin{tabular}[c]{@{}c@{}}Citation \\ Rate\end{tabular} & ROUGE-1 & ROUGE-L & \begin{tabular}[c]{@{}c@{}}Entity \\ Recall\end{tabular} & \begin{tabular}[c]{@{}c@{}}Numerical \\ Recall\end{tabular}  \\
        \midrule
        RAG & 14.03 (9.89) & 2114.42 & 103.28 & 7.51 & 82.29 & 70.92 & 92.92 & 47.43 & 16.08 & 16.18 & 25.48 \\
        STORM & 9.80 (8.75) & 1917.26 & 70.08 & 6.01 & 77.49 & 75.71 & 90.65 & 43.51 & 15.50 & 13.37 & 24.22 \\
        MOG (ours)& \textbf{24.86$^{\dagger}$} (8.72)& \textbf{2490.03$^{\dagger}$} & \textbf{155.25$^{\dagger}$} & \textbf{13.87$^{\dagger}$} & \textbf{85.07} & \textbf{80.05$^{\dagger}$} & \textbf{94.84} & \textbf{50.38$^{\dagger}$} & \textbf{16.46$^{\dagger}$} & \textbf{18.96$^{\dagger}$} & \textbf{35.99$^{\dagger}$} \\
        \midrule
        Human-written & 15.89 (5.78)& 2429.53 & 219.97 & 19.67 & - & - & - & 100.00 & 100.00 & 100.00 & 100.00 \\
        \bottomrule
    \end{tabular}
    }
    \makebox[\textwidth]{\small (b) Results on WikiStart.}
    \resizebox{\textwidth}{!}{
        \Large
        \begin{tabular}{l||*{4}{c}||*{3}{c}||*{4}{c}}
        \toprule
        & \multicolumn{4}{c||}{\textbf{Informativeness}} & \multicolumn{3}{c||}{\textbf{Verifiability}} & \multicolumn{4}{c}{\textbf{Comparison with Human-written Articles}}  \\
        & \begin{tabular}[c]{@{}c@{}}Section \\ Count\end{tabular} & \begin{tabular}[c]{@{}c@{}}Word \\ Count\end{tabular} & \begin{tabular}[c]{@{}c@{}}Entity \\ Count\end{tabular} & \begin{tabular}[c]{@{}c@{}}Numerical \\ Count\end{tabular} & \begin{tabular}[c]{@{}c@{}}Citation \\ Recall\end{tabular} & \begin{tabular}[c]{@{}c@{}}Citation \\ Precision\end{tabular} & \begin{tabular}[c]{@{}c@{}}Citation \\ Rate\end{tabular} & ROUGE-1 & ROUGE-L & \begin{tabular}[c]{@{}c@{}}Entity \\ Recall\end{tabular} & \begin{tabular}[c]{@{}c@{}}Numerical \\ Recall\end{tabular}  \\
        \midrule
        RAG & 12.26 (9.43) & 1674.16 & 73.52 & 5.45 & 72.66 & 63.22 & 93.55 & 60.56 & 37.58 & 26.15 & 59.73 \\
        STORM & 8.97 (8.39) & 1646.67 & 52.00 & 4.98 & 67.72 & 66.50 & 86.09 & 59.36 & 37.69 & 20.06 & 59.46 \\
        MOG (ours)& \textbf{21.23$^{\dagger}$} (7.95) & \textbf{2048.81$^{\dagger}$} & \textbf{131.40$^{\dagger}$} & \textbf{12.57$^{\dagger}$} & \textbf{82.14$^{\dagger}$} & \textbf{76.57$^{\dagger}$} & \textbf{95.91} & \textbf{67.29$^{\dagger}$} & \textbf{40.97$^{\dagger}$} & \textbf{31.33$^{\dagger}$} & \textbf{70.18$^{\dagger}$} \\
        \midrule
        Human-written & 5.18 (3.20) & 249.09 & 25.07 & 2.64 & - & - & - & 100.00 & 100.00 & 100.00 & 100.00 \\
        \bottomrule
    \end{tabular}
    }
    \caption{Results evaluated on the high-resource FreshWiki dataset and low-resource WikiStart dataset. The section count represents all sections and subsections, with numbers in parentheses indicating only first-level sections. Citation results for human-written articles are omitted due to some links being inaccessible. $\dagger$ denotes significant differences ($p < 0.05$) from a paired $t$-test between \textbf{MOG} and the best baseline.}
    \label{tab:main_results}
\end{table*}

\subsection{Baselines}
In this study, we employ two main baselines: RAG and STORM~\citep{shao-etal-2024-assisting}. While extensive research exists on automating Wikipedia article generation, these methodologies align well with our task definition, which aims to produce full-length Wikipedia articles on a broad range of topics.

\begin{enumerate}
    \item \textit{RAG} operates through a multi-step retrieval process. It first conducts a web search for topic-relevant documents to create an outline. Then, it performs additional searches by combining the original topic with each section heading. Finally, it generates section contents using the most relevant document chunks as working context.
    \item \textit{STORM} focuses on the pre-writing phase. It engages in multi-perspective conversations to gather comprehensive information and subsequently creates a detailed outline from the conversation history. It then follows the same content generation process as \textit{RAG}.
\end{enumerate}

\subsection{Automatic Metrics}
Our work aims to enhance the informativeness and verifiability of the generated articles while maintaining fluency and coherence. To assess informativeness, we report statistical measures including article length, entity count~\citep{Honnibal_spaCy_Industrial-strength_Natural_2020}, and numerical data count (e.g., dates, statistics). For verifiability, we adopt Citation Recall and Citation Precision as defined by \citet{gao2023enabling}. Citation Recall reflects the extent to which the generated sentences can be supported by the citations, while Citation Precision indicates the ability to avoid overciting unnecessary information. Additionally, we introduce the Citation Rate, which measures the proportion of generated sentences that include a citation. To assess whether a citation properly supports the corresponding sentence, the evaluation is conducted using \texttt{gpt-4o-mini-2024-07-18}. For comparison with human-written articles, we report recall metrics for ROUGE-1, ROUGE-L~\citep{lin2004rouge}, entities, and numerical data. In evaluating Wikipedia generation, we prioritize recall over precision or F-measure, as there is no definitive version of an article that can be considered the ground truth. Given the open-ended and informational nature of Wikipedia articles, our goal is to ensure that generated content is comprehensive without penalizing it for including more information than human-written counterparts.

Additionally, we incorporate LLM-based metrics~\citep{shao-etal-2024-assisting, zhang2024bench} to evaluate aspects of article quality that traditional metrics may overlook. These include the \textit{Interest Score}, \textit{Organization Score}, and \textit{Focus Score}, assigned by Prometheus-2~\citep{kim2024prometheus} using a 5-point rubric.\footnote{\url{https://huggingface.co/prometheus-eval/prometheus-7b-v2.0}} The Interest Score measures how engaging and thought-provoking the article is. The Organization Score evaluates the structural quality, ensuring the article follows Wikipedia guidelines with clear sections, logical flow, and appropriate headings. The Focus Score assesses whether the article remains on topic, maintaining relevance to the core subject and avoiding unnecessary digressions.


\begin{figure}[t]
    \centering
    \begin{subfigure}[b]{\columnwidth}
        \centering
        \includegraphics[width=\columnwidth]{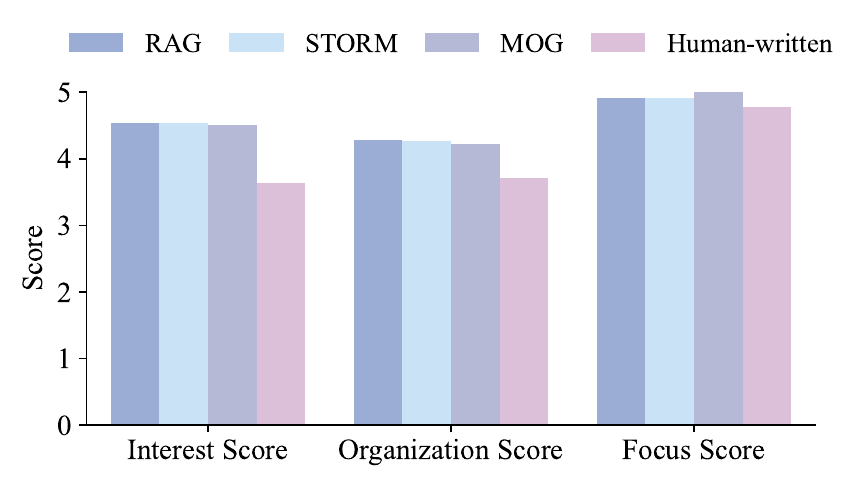}
        \caption{FreshWiki Dataset}
        \label{fig:llm_freshwiki}
    \end{subfigure}
    \hfill
    \begin{subfigure}[b]{\columnwidth}
        \centering
        \includegraphics[width=\columnwidth]{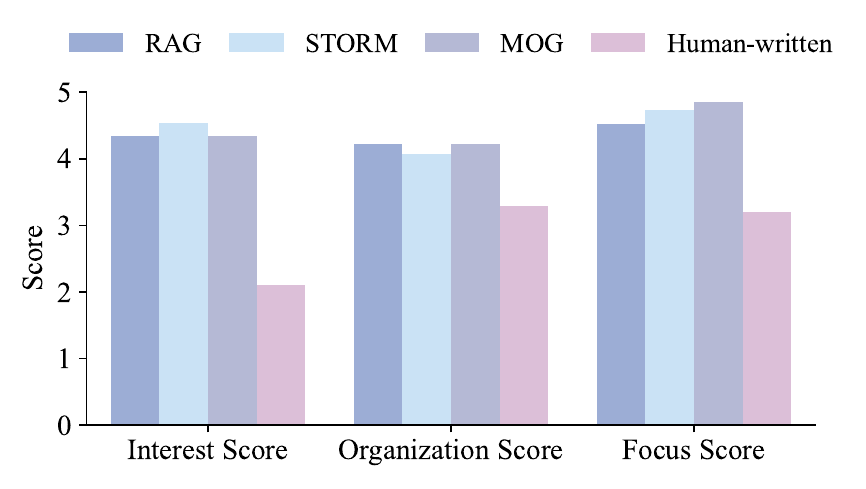}
        \caption{WikiStart Dataset}
        \label{fig:llm_WikiStart}
    \end{subfigure}
    \caption{Results of LLM-based metrics on different datasets. The article length is capped at 10,000 words, adhering to the truncation method proposed by~\citet{shao-etal-2024-assisting}.}
    %
    \label{fig:llm_results}
\end{figure}
\section{Experiments}
We use the DSPy framework~\citep{khattab2023dspy} to prompt the LLMs used in this study (see Appendix~\ref{appendix:implementation}). Specifically, \texttt{gpt-4o-2024-08-06} is employed for outline planning and article generation, while \texttt{gpt-4o-mini-2024-07-18} handles all other LLM tasks. All LLM calls are configured with a temperature of 0 and a \texttt{top\_p} value of 1. For query searching, URLs are retrieved via VALUESERP\footnote{\url{https://docs.trajectdata.com/valueserp/search-api/overview}}, with content scraped manually from the corresponding webpages. In contrast, the STORM baseline retrieves web content from the You.com search API\footnote{\url{https://documentation.you.com/api-reference/search}} using curated queries.

We fine-tuned the \texttt{all-miniLM-L6-v2} model\footnote{\url{https://huggingface.co/sentence-transformers/all-MiniLM-L6-v2}} as the encoder for the vector database, utilizing section headings and their corresponding sentences from Wikipedia as training data~(Appendix \ref{appendix:implementation_finetune}). This fine-tuning procedure aligns the vector space representations of section headings and memory units, improving the accuracy of clustering memory units by section and enhancing the reliability of similarity calculations between section headings and memory units.

\subsection{Main Results}
Table~\ref{tab:main_results} summarizes the performance of various methods on the FreshWiki and WikiStart datasets. We report reference-free metrics along two dimensions: informativeness and verifiability. Furthermore, we compare the generated articles to human-written ones. Notably, the human-written articles in the WikiStart dataset are considerably shorter than those in the FreshWiki dataset, with an average length of approximately 249 words. Additionally, Figure~\ref{fig:llm_results} presents LLM-based metrics, which show that MOG performs commendably in other dimensions beyond just informativeness and verifiability. 

\vspace{1em}
\noindent\textbf{MOG Outperforms Baselines in Informativeness Across Datasets.}\quad
On both the FreshWiki and WikiStart dataset, MOG generates content with higher word and entity counts, indicating more comprehensive coverage. MOG expands sections into approximately three times as many subsections, mirroring trends observed in human-written articles despite having the fewest first-level sections. This expansion is not merely superficial; the improved recall of n-grams, entities, and numerical data demonstrates that MOG provides more substantial and relevant information. Moreover, results show that MOG achieves higher information density; on WikiStart, for example, it increases entity count by 79\% over RAG and 153\% over STORM, with only a 22–24\% increase in length—indicating the gains are not merely due to longer outputs.



\vspace{1em}
\noindent\textbf{MOG outperforms baselines in generating verifiable content.}\quad
On both FreshWiki and WikiStart, MOG outperforms other methods in Citation Recall, Citation Precision, and Citation Rate. This suggests that MOG not only generates more content, but also produces content that is more verifiable. Among these metrics, we prioritize Citation Recall, as it ensures that the generated content is based on reliable sources and not fabricated. Notably, MOG achieves higher Citation Recall despite having to cite from a larger pool of fine-grained memory units, each containing less information than the larger document chunks used by RAG and STORM. This strong performance enables MOG to integrate information from up to 100 sources, compared to just 5 sources in the case of RAG and STORM given the same working context length. Additionally, the use of smaller memory units enhances verifiability by directly linking the generated content to the specific pieces of corresponding information of the sources, making it easier for users to verify individual claims without having to sift through lengthy documents.

\vspace{1em}
\noindent\textbf{Comparison of method performance in high- and low-resource scenarios.}\quad
All methods show a decrease in informativeness when transitioning from the FreshWiki dataset to the WikiStart dataset, reflecting the fact that topics in WikiStart generally have less available online information than those in FreshWiki. In the low-resource scenario, both RAG and STORM experience an approximately 10\% drop in Citation Recall, while MOG only sees a 2.93\% decline. This suggests that MOG is more robust in low-resource environments, which is consistent with real-world applications where such conditions are common. Additionally, both RAG and MOG, which utilize a post-hoc citation approach, show an increase in Citation Rates on the WikiStart dataset. In contrast, STORM, which generates citations in real time during content creation, experiences a 4.56\% decrease. This underscores the advantages of separating the citation process given the current limitations of LLMs.

\vspace{1em} \noindent\textbf{LLM-based metrics.}\quad MOG achieves the highest Focus Score, with a perfect 5 on the FreshWiki dataset, and performs comparably to baselines on Organization and Interest. These results suggest that MOG maintains strong writing quality despite not being explicitly optimized for style. While the improvements in LLM-based scores are relatively modest, they indicate that MOG’s gains in informativeness and verifiability are not achieved at the expense of fluency or coherence. Notably, human-written articles receive low Interest and Organization scores, likely due to being unprocessed crawled content with markup, broken text, and tables. These scores improve notably after light formatting, while Focus remains stable (see Appendix~\ref{appendix:human_written_article}).


\begin{table}[t]
\centering

\makebox[\columnwidth]{\small (a) Results on FreshWiki.} 
\resizebox{0.95\columnwidth}{!}{
    \centering
    \begin{tabular}{c||c|c|c}
        \toprule
         & \shortstack{Total Webpages \\ Collected} & \shortstack{Total Webpages \\ Cited} & \shortstack{Utilization \\ Rate} \\
        \toprule
        RAG & \textbf{76.61} & 23.00 & 31.09 \\
        MOG & 65.28 & \textbf{49.21} & \textbf{77.67} \\
        \bottomrule
    \end{tabular}
}

\vspace{0.5em}

\makebox[\columnwidth]{\small (b) Results on WikiStart.} 
\resizebox{0.95\columnwidth}{!}{
    \centering
    \begin{tabular}{c||c|c|c}
        \toprule
        & \shortstack{Total Webpages \\ Collected} & \shortstack{Total Webpages \\ Cited} & \shortstack{Utilization \\ Rate} \\
        \toprule
        RAG & \textbf{65.17} & 17.60 & 28.34 \\
        MOG & 48.57 & \textbf{36.33} & \textbf{75.44} \\
        \bottomrule
    \end{tabular}
}

\caption{Comparison of webpage collection and utilization rates Between RAG and MOG. RAG stores entire document chunks as memory, while MOG extracts relevant factoids as memory units.}
\label{tab:info_analysis}
\end{table}

\begin{table}[t]
\centering
\resizebox{0.95\columnwidth}{!}{
    \begin{tabular}{l||c|c|c|c}
        \toprule
         & \shortstack{Section \\ Count} & \shortstack{Word \\ Count} & \shortstack{Entity \\ Count} & \shortstack{Webpages \\ Count} \\
        \toprule
        MOG & \textbf{21.46} (7.78)& \textbf{2046.28} & \textbf{141.26} & \textbf{53.98} \\
        ~-w/o SE & 9.78 (5.32)& 1041.96 & 83.66 & 18.30 \\
        ~-w/o MO & 8.18 (8.18)& 1360.04 & 105.28 & 55.3 \\
        \bottomrule
    \end{tabular}
}
\caption{Ablation study results: "w/o SE" denotes the exclusion of the Subtopic Explorer, while "w/o MO" refers to the exclusion of the Memory Organization procedure. For the complete results, refer to Table~\ref{tab:full_ablation_studies} in the Appendix.}

\label{tab:ablation}
\end{table}

\section{Further Analysis}

\subsection{Information Acquisition and Utilization}

Table~\ref{tab:info_analysis} presents a comparison of the information acquisition and utilization strategies employed by RAG and MOG across the FreshWiki and WikiStart datasets. The FreshWiki dataset comprises a greater number of collected webpages than WikiStart, highlighting the relative scarcity of resources for topics in WikiStart. Despite using more search operations during memory construction, MOG achieves greater efficiency than RAG by selectively extracting only webpages directly relevant to the topic. This approach naturally filters unrelated information and results in fewer retained webpages. By focusing exclusively on relevant material, MOG reduces the likelihood of inaccuracies in the generated articles. Furthermore, our analysis reveals that MOG’s fine-grained memory units achieve a higher utilization rate than the document chunks used by RAG. Specifically, over 75\% of MOG's memory units contribute to article generation, compared to only about 30\% in RAG, highlighting MOG's superior efficiency in processing and leveraging acquired information.

\subsection{Ablation Studies}
We conducted ablation studies to isolate the influence of the subtopic explorer and memory organization components on the MOG procedure. The study was conducted on 50 articles, with 25 sampled from the FreshWiki dataset and 25 from the WikiStart dataset. As shown in Table~\ref{tab:ablation}, removing the subtopic explorer and memory organization components leads to a reduction in the entity count by 40.78\% and 25.47\% respectively. Furthermore, the subtopic explorer plays a critical role in identifying and retrieving valid, topic-relevant webpages, as its absence reduces the number of such webpages by approximately threefold. Even with the reduced memory content caused by the removal of the subtopic explorer, the memory organization component proves to be effective: MOG without the subtopic explorer produces 9.78 sections on average, compared to 8.18 sections generated when the memory organization component is removed. 


\section{Conclusion}
The proposed MOG framework represents a novel and effective approach to the autonomous generation of Wikipedia articles. By leveraging a hierarchical memory architecture, MOG addresses key challenges in informativeness, verifiability, and hallucination reduction in long-form text generation. Experimental results on high-resource (FreshWiki) and low-resource (WikiStart) datasets demonstrate that MOG consistently outperforms baselines such as RAG and STORM. Notably, MOG's recursive memory organization enables the dynamic adjustment of section depth to maximize memory utilization. MOG's effectiveness highlights its potential for broader applications in long-form structured text generation. Furthermore, the framework offers the potential to overcome the limitations of existing training data by facilitating the creation of high-quality, citation-rich datasets with complete and accessible inputs. These datasets may enhance the accuracy, coherence, and reliability of models, particularly in long-context scenarios.
\section*{Limitations}
Our work enhances the informativeness and verifiability of generated Wikipedia articles; however, several challenges remain. Conflicting information from diverse sources requires more robust mechanisms for fact-checking and source validation. The system must also be able to counteract intentional misinformation. Additionally, relying on factoids as the primary memory unit may result in the omission of critical temporal or sequential details, impacting domains where narrative order is crucial, such as sporting event recaps. While MOG's effectiveness with open-sourced models is empirically verified in production settings, further experiments are necessary to explore its full potential and limitations. Future research will examine the applicability of MOG to other domains like financial reporting to understand where it excels in long-form text generation.

\section*{Ethics Statement}

This research complies with the ACL Ethics Policy. To the best of our knowledge, it does not raise any ethical concerns. We have carefully examined the potential ethical implications of our work, including its methodology, applications, and broader impacts, and have not identified any issues that would pose ethical risks.


\section*{Acknowledgement}
We thank the anonymous reviewers for their helpful comments on this paper. 
This work was partially supported by National Natural Science Foundation of China (No. 92470205).

\bibliography{main}

\begin{thebibliography}{30}
\providecommand{\natexlab}[1]{#1}

\bibitem[{Achiam et~al.(2023)Achiam, Adler, Agarwal, Ahmad, Akkaya, Aleman, Almeida, Altenschmidt, Altman, Anadkat et~al.}]{achiam2023gpt}
Josh Achiam, Steven Adler, Sandhini Agarwal, Lama Ahmad, Ilge Akkaya, Florencia~Leoni Aleman, Diogo Almeida, Janko Altenschmidt, Sam Altman, Shyamal Anadkat, et~al. 2023.
\newblock Gpt-4 technical report.
\newblock \emph{arXiv preprint arXiv:2303.08774}.

\bibitem[{Banerjee and Mitra(2015)}]{banerjee-mitra-2015-wikikreator}
Siddhartha Banerjee and Prasenjit Mitra. 2015.
\newblock \href {https://doi.org/10.3115/v1/P15-1084} {{W}iki{K}reator: Improving {W}ikipedia stubs automatically}.
\newblock In \emph{Proceedings of the 53rd Annual Meeting of the Association for Computational Linguistics and the 7th International Joint Conference on Natural Language Processing (Volume 1: Long Papers)}, pages 867--877, Beijing, China. Association for Computational Linguistics.

\bibitem[{Bird(2006)}]{bird2006nltk}
Steven Bird. 2006.
\newblock Nltk: the natural language toolkit.
\newblock In \emph{Proceedings of the COLING/ACL 2006 Interactive Presentation Sessions}, pages 69--72.

\bibitem[{Brown(2020)}]{brown2020language}
Tom~B Brown. 2020.
\newblock Language models are few-shot learners.
\newblock \emph{arXiv preprint arXiv:2005.14165}.

\bibitem[{Chen et~al.(2023)Chen, Wang, Chen, Yu, Ma, Zhao, Zhang, and Yu}]{chen2023dense}
Tong Chen, Hongwei Wang, Sihao Chen, Wenhao Yu, Kaixin Ma, Xinran Zhao, Hongming Zhang, and Dong Yu. 2023.
\newblock Dense x retrieval: What retrieval granularity should we use?
\newblock \emph{arXiv preprint arXiv:2312.06648}.

\bibitem[{Fan and Gardent(2022)}]{fan-gardent-2022-generating}
Angela Fan and Claire Gardent. 2022.
\newblock \href {https://doi.org/10.18653/v1/2022.acl-long.586} {Generating biographies on {W}ikipedia: The impact of gender bias on the retrieval-based generation of women biographies}.
\newblock In \emph{Proceedings of the 60th Annual Meeting of the Association for Computational Linguistics (Volume 1: Long Papers)}, pages 8561--8576, Dublin, Ireland. Association for Computational Linguistics.

\bibitem[{Gao et~al.(2024)Gao, Jiang, Yang, Zeng, Lu, Blum, She, Jiang, and Li}]{gao2024evaluating}
Fan Gao, Hang Jiang, Rui Yang, Qingcheng Zeng, Jinghui Lu, Moritz Blum, Tianwei She, Yuang Jiang, and Irene Li. 2024.
\newblock Evaluating large language models on wikipedia-style survey generation.
\newblock In \emph{Findings of the Association for Computational Linguistics ACL 2024}, pages 5405--5418.

\bibitem[{Gao et~al.(2023{\natexlab{a}})Gao, Yen, Yu, and Chen}]{gao2023enabling}
Tianyu Gao, Howard Yen, Jiatong Yu, and Danqi Chen. 2023{\natexlab{a}}.
\newblock Enabling large language models to generate text with citations.
\newblock \emph{arXiv preprint arXiv:2305.14627}.

\bibitem[{Gao et~al.(2023{\natexlab{b}})Gao, Xiong, Gao, Jia, Pan, Bi, Dai, Sun, and Wang}]{gao2023retrieval}
Yunfan Gao, Yun Xiong, Xinyu Gao, Kangxiang Jia, Jinliu Pan, Yuxi Bi, Yi~Dai, Jiawei Sun, and Haofen Wang. 2023{\natexlab{b}}.
\newblock Retrieval-augmented generation for large language models: A survey.
\newblock \emph{arXiv preprint arXiv:2312.10997}.

\bibitem[{Guu et~al.(2020)Guu, Lee, Tung, Pasupat, and Chang}]{guu2020retrieval}
Kelvin Guu, Kenton Lee, Zora Tung, Panupong Pasupat, and Mingwei Chang. 2020.
\newblock Retrieval augmented language model pre-training.
\newblock In \emph{International conference on machine learning}, pages 3929--3938. PMLR.

\bibitem[{Honnibal et~al.(2020)Honnibal, Montani, Van~Landeghem, and Boyd}]{Honnibal_spaCy_Industrial-strength_Natural_2020}
Matthew Honnibal, Ines Montani, Sofie Van~Landeghem, and Adriane Boyd. 2020.
\newblock \href {https://doi.org/10.5281/zenodo.1212303} {{spaCy: Industrial-strength Natural Language Processing in Python}}.

\bibitem[{Hu et~al.(2023)Hu, Fu, Du, Luo, Zhao, and Zhao}]{hu2023chatdb}
Chenxu Hu, Jie Fu, Chenzhuang Du, Simian Luo, Junbo Zhao, and Hang Zhao. 2023.
\newblock Chatdb: Augmenting llms with databases as their symbolic memory.
\newblock \emph{arXiv preprint arXiv:2306.03901}.

\bibitem[{Karpukhin et~al.(2020)Karpukhin, O{\u{g}}uz, Min, Lewis, Wu, Edunov, Chen, and Yih}]{karpukhin2020dense}
Vladimir Karpukhin, Barlas O{\u{g}}uz, Sewon Min, Patrick Lewis, Ledell Wu, Sergey Edunov, Danqi Chen, and Wen-tau Yih. 2020.
\newblock Dense passage retrieval for open-domain question answering.
\newblock \emph{arXiv preprint arXiv:2004.04906}.

\bibitem[{Khattab et~al.(2023)Khattab, Singhvi, Maheshwari, Zhang, Santhanam, Vardhamanan, Haq, Sharma, Joshi, Moazam et~al.}]{khattab2023dspy}
Omar Khattab, Arnav Singhvi, Paridhi Maheshwari, Zhiyuan Zhang, Keshav Santhanam, Sri Vardhamanan, Saiful Haq, Ashutosh Sharma, Thomas~T Joshi, Hanna Moazam, et~al. 2023.
\newblock Dspy: Compiling declarative language model calls into self-improving pipelines.
\newblock \emph{arXiv preprint arXiv:2310.03714}.

\bibitem[{Kim et~al.(2023)Kim, Shin, Cho, Jang, Longpre, Lee, Yun, Shin, Kim, Thorne et~al.}]{kim2023prometheus}
Seungone Kim, Jamin Shin, Yejin Cho, Joel Jang, Shayne Longpre, Hwaran Lee, Sangdoo Yun, Seongjin Shin, Sungdong Kim, James Thorne, et~al. 2023.
\newblock Prometheus: Inducing evaluation capability in language models.
\newblock In \emph{NeurIPS 2023 Workshop on Instruction Tuning and Instruction Following}.

\bibitem[{Kim et~al.(2024)Kim, Suk, Longpre, Lin, Shin, Welleck, Neubig, Lee, Lee, and Seo}]{kim2024prometheus}
Seungone Kim, Juyoung Suk, Shayne Longpre, Bill~Yuchen Lin, Jamin Shin, Sean Welleck, Graham Neubig, Moontae Lee, Kyungjae Lee, and Minjoon Seo. 2024.
\newblock Prometheus 2: An open source language model specialized in evaluating other language models.
\newblock \emph{arXiv preprint arXiv:2405.01535}.

\bibitem[{Kojima et~al.(2022)Kojima, Gu, Reid, Matsuo, and Iwasawa}]{kojima2022large}
Takeshi Kojima, Shixiang~Shane Gu, Machel Reid, Yutaka Matsuo, and Yusuke Iwasawa. 2022.
\newblock Large language models are zero-shot reasoners.
\newblock \emph{Advances in neural information processing systems}, 35:22199--22213.

\bibitem[{Li et~al.(2024)Li, Qin, Xiao, Chen, Luo, Shao, Lian, and Liu}]{li2024makingtextembeddersfewshot}
Chaofan Li, MingHao Qin, Shitao Xiao, Jianlyu Chen, Kun Luo, Yingxia Shao, Defu Lian, and Zheng Liu. 2024.
\newblock \href {https://arxiv.org/abs/2409.15700} {Making text embedders few-shot learners}.
\newblock \emph{Preprint}, arXiv:2409.15700.

\bibitem[{Lin(2004)}]{lin2004rouge}
Chin-Yew Lin. 2004.
\newblock Rouge: A package for automatic evaluation of summaries.
\newblock In \emph{Text summarization branches out}, pages 74--81.

\bibitem[{Liu et~al.(2018)Liu, Saleh, Pot, Goodrich, Sepassi, Kaiser, and Shazeer}]{liu2018generating}
Peter~J Liu, Mohammad Saleh, Etienne Pot, Ben Goodrich, Ryan Sepassi, Lukasz Kaiser, and Noam Shazeer. 2018.
\newblock Generating wikipedia by summarizing long sequences.
\newblock \emph{arXiv preprint arXiv:1801.10198}.

\bibitem[{Modarressi et~al.(2023)Modarressi, Imani, Fayyaz, and Sch{\"u}tze}]{modarressi2023ret}
Ali Modarressi, Ayyoob Imani, Mohsen Fayyaz, and Hinrich Sch{\"u}tze. 2023.
\newblock Ret-llm: Towards a general read-write memory for large language models.
\newblock \emph{arXiv preprint arXiv:2305.14322}.

\bibitem[{Park et~al.(2023)Park, O'Brien, Cai, Morris, Liang, and Bernstein}]{park2023generative}
Joon~Sung Park, Joseph O'Brien, Carrie~Jun Cai, Meredith~Ringel Morris, Percy Liang, and Michael~S Bernstein. 2023.
\newblock Generative agents: Interactive simulacra of human behavior.
\newblock In \emph{Proceedings of the 36th annual acm symposium on user interface software and technology}, pages 1--22.

\bibitem[{Qian et~al.(2023)Qian, Zhu, Dou, Gu, Zhang, Liu, Lai, Cao, Nie, and Wen}]{qian2023webbrain}
Hongjing Qian, Yutao Zhu, Zhicheng Dou, Haoqi Gu, Xinyu Zhang, Zheng Liu, Ruofei Lai, Zhao Cao, Jian-Yun Nie, and Ji-Rong Wen. 2023.
\newblock Webbrain: Learning to generate factually correct articles for queries by grounding on large web corpus.
\newblock \emph{arXiv preprint arXiv:2304.04358}.

\bibitem[{Reimers and Gurevych(2019)}]{reimers-2019-sentence-bert}
Nils Reimers and Iryna Gurevych. 2019.
\newblock \href {https://arxiv.org/abs/1908.10084} {Sentence-bert: Sentence embeddings using siamese bert-networks}.
\newblock In \emph{Proceedings of the 2019 Conference on Empirical Methods in Natural Language Processing}. Association for Computational Linguistics.

\bibitem[{Sanderson and Croft(1999)}]{sanderson1999deriving}
Mark Sanderson and Bruce Croft. 1999.
\newblock Deriving concept hierarchies from text.
\newblock In \emph{Proceedings of the 22nd annual international ACM SIGIR conference on Research and development in information retrieval}, pages 206--213.

\bibitem[{Sauper and Barzilay(2009)}]{sauper-barzilay-2009-automatically}
Christina Sauper and Regina Barzilay. 2009.
\newblock \href {https://aclanthology.org/P09-1024} {Automatically generating {W}ikipedia articles: A structure-aware approach}.
\newblock In \emph{Proceedings of the Joint Conference of the 47th Annual Meeting of the {ACL} and the 4th International Joint Conference on Natural Language Processing of the {AFNLP}}, pages 208--216, Suntec, Singapore. Association for Computational Linguistics.

\bibitem[{Shao et~al.(2024)Shao, Jiang, Kanell, Xu, Khattab, and Lam}]{shao-etal-2024-assisting}
Yijia Shao, Yucheng Jiang, Theodore Kanell, Peter Xu, Omar Khattab, and Monica Lam. 2024.
\newblock \href {https://doi.org/10.18653/v1/2024.naacl-long.347} {Assisting in writing {W}ikipedia-like articles from scratch with large language models}.
\newblock In \emph{Proceedings of the 2024 Conference of the North American Chapter of the Association for Computational Linguistics: Human Language Technologies (Volume 1: Long Papers)}, pages 6252--6278, Mexico City, Mexico. Association for Computational Linguistics.

\bibitem[{Wang et~al.(2024)Wang, Yu, Wang, Chen, Zhu, and Dou}]{wang2024richrag}
Shuting Wang, Xin Yu, Mang Wang, Weipeng Chen, Yutao Zhu, and Zhicheng Dou. 2024.
\newblock Richrag: Crafting rich responses for multi-faceted queries in retrieval-augmented generation.
\newblock \emph{arXiv preprint arXiv:2406.12566}.

\bibitem[{Zhang et~al.(2024{\natexlab{a}})Zhang, Bai, Lv, Gu, Liu, Zou, Cao, Hou, Dong, Feng et~al.}]{zhang2024longcite}
Jiajie Zhang, Yushi Bai, Xin Lv, Wanjun Gu, Danqing Liu, Minhao Zou, Shulin Cao, Lei Hou, Yuxiao Dong, Ling Feng, et~al. 2024{\natexlab{a}}.
\newblock Longcite: Enabling llms to generate fine-grained citations in long-context qa.
\newblock \emph{arXiv preprint arXiv:2409.02897}.

\bibitem[{Zhang et~al.(2024{\natexlab{b}})Zhang, Chen, Hu, Xu, Chen, Hao, Han, Thai, Wang, Liu et~al.}]{zhang2024bench}
Xinrong Zhang, Yingfa Chen, Shengding Hu, Zihang Xu, Junhao Chen, Moo Hao, Xu~Han, Zhen Thai, Shuo Wang, Zhiyuan Liu, et~al. 2024{\natexlab{b}}.
\newblock ∞ bench: Extending long context evaluation beyond 100k tokens.
\newblock In \emph{Proceedings of the 62nd Annual Meeting of the Association for Computational Linguistics (Volume 1: Long Papers)}, pages 15262--15277.

\end{thebibliography}
\clearpage
\appendix
\section{Dataset Details}
\label{appendix:dataset}
To construct the WikiStart dataset, we utilized PetScan\footnote{\url{https://petscan.wmcloud.org/}} to filter and curate relevant Wikipedia entries. Our selection criteria included articles categorized under ``All articles to be expanded''\footnote{\url{https://en.wikipedia.org/wiki/Category:Articles_to_be_expanded}} with a last edit date prior to January 1, 2023. Most articles in this category are rated as Start-class\footnote{\url{https://en.wikipedia.org/wiki/Category:Start-Class_articles}}, which are slightly more informative than Stub articles but still require major expansion. We chose this category to ensure that sufficient information about the topics is available online. We further refined the dataset by selecting articles whose lengths ranged from 3,000 to 20,000 bytes. To ensure the exclusion of list-based entries, we omitted topics with titles beginning with ``List of ...'', as these typically consist of lists rather than standard narrative articles. This filtering process resulted in a total of 1,246 articles.

From this pool, we randomly selected 100 topics to form the evaluation dataset, aligning our methodology with the standards established by FreshWiki. For each selected topic, we manually scraped the corresponding human-written articles from Wikipedia using the \texttt{wikipediaapi} library\footnote{\url{https://pypi.org/project/Wikipedia-API/}}. Importantly, we retained tabular and list data within the articles to enhance the reliability of our informativeness metrics, including entity and numerical recall. This approach ensures that the dataset not only reflects comprehensive article content but also maintains the structural integrity necessary for accurate evaluation.


\begin{table}[t]
\centering
\resizebox{0.95\columnwidth}{!}{
    \begin{tabular}{l||c|c}
        \toprule
         Model & \# Parameter & Precision \\
        \toprule
        \texttt{all-MiniLM-L6-v2}\footnotemark[14] & 22M & 50.96 \\
        \texttt{dpr-single-nq-base}\footnotemark[15] & 110M & 48.21 \\
         \texttt{dpr-multiset-base}\footnotemark[16] & 110M & 47.68 \\
         \texttt{bge-en-icl}\footnotemark[17] & 7.11B & 52.07 \\
        \bottomrule
    \end{tabular}
}
\caption{Preliminary analysis of classification precision of section heading and sentences of different encoders.}
\label{tab:encoder_precision}
\end{table}

\section{Implementation Details}
We provide the complete set of prompts used in MOG in Listing~\ref{lst:mog-prompts} and Listing~\ref{lst:mog-prompts-continue}.

\subsection{Fine-tuning of the Encoder}
\label{appendix:implementation_finetune}
We began our work using the July 20, 2024 English Wikipedia dump, which contains approximately 11 million articles. We parsed each article into a hierarchical structure of sections and sentences, ensuring that every section included all sentences from its nested subsections. Here, the sentence serves as an approximation of the memory unit defined in Section~\ref{sec:method}. From this dataset, we randomly selected a small subset of section–sentence pairs as a preliminary evaluation set.

\begin{equation}
l^* = \arg\max_l \text{sim}(e, l)
\label{eq:optimal_section}
\end{equation}

To evaluate the encoder's performance, we examined its effectiveness in assigning each sentence \( e \) to the most semantically appropriate section heading \( l^* \) drawn from the siblings of its parent section \(\{l\}\), as indicated by Equation~\ref{eq:optimal_section}. For example, a sentence appearing under Section~1.1 is compared only against headings such as Section~1.2, Section~1.3, and so forth. Table~\ref{tab:encoder_precision} reports the results for several models, including \texttt{all-MiniLM-L6-v2}, \texttt{dpr-single-nq-base}~\citep{reimers-2019-sentence-bert}, \texttt{dpr-multiset-base}~\citep{karpukhin2020dense}, and \texttt{bge-en-icl}~\citep{li2024makingtextembeddersfewshot}. The \texttt{all-MiniLM-L6-v2} model, which employs the same encoder for both section headings and sentences, outperformed the dual-encoder DPR variants. Although \texttt{bge-en-icl} achieved a slightly higher accuracy (1.11\% improvement over \texttt{all-MiniLM-L6-v2}), this gain comes at the cost of an approximately 300-fold increase in parameter count. We therefore selected \texttt{all-MiniLM-L6-v2} as our base model for subsequent fine-tuning.

\footnotetext[14]{\url{https://huggingface.co/sentence-transformers/all-MiniLM-L6-v2}}
\footnotetext[15]{\url{https://huggingface.co/facebook/dpr-question_encoder-single-nq-base}}
\footnotetext[16]{\url{https://huggingface.co/facebook/dpr-question_encoder-multiset-base}}
\footnotetext[17]{\url{https://huggingface.co/BAAI/bge-en-icl}}

\begin{figure}[t]
    \centering
    \includegraphics[width=0.95\columnwidth]{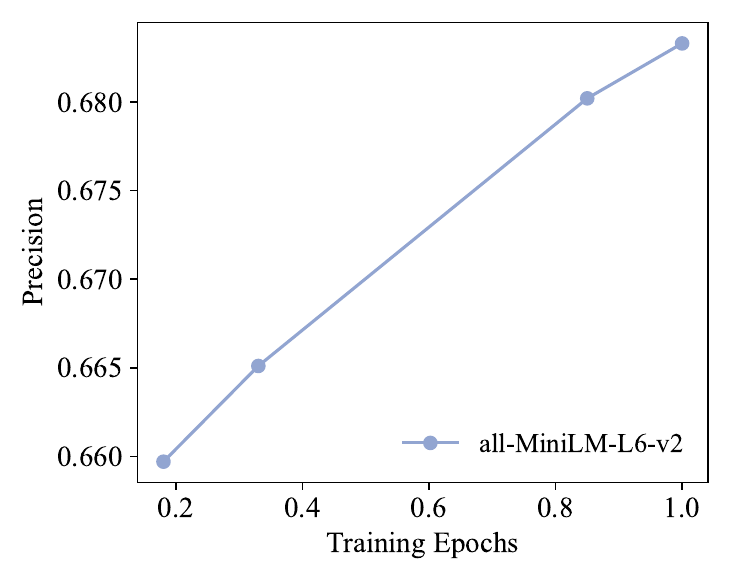}
    \caption{Precision achieved by the encoder during fine-tuning across training epochs.}
    \label{fig:encoder_training}
\end{figure}

To further refine the model, we employ a contrastive learning approach. Specifically, for each section, sentences from sibling sections are used as hard negative samples. This method is intended to sharpen the model’s capacity to distinguish between closely related headings and thereby improve its identification of the most semantically relevant heading for a given sentence. As illustrated in Figure~\ref{fig:encoder_training}, this approach yields a 17.37\% improvement in precision after a single epoch of training. Subsequent epochs did not substantially increase performance, and training was halted due to computational limitations. Although these findings are preliminary, scaling the model and extending training may offer further improvements.

\begin{figure}[t]
    \centering
    \includegraphics[width=0.95\columnwidth]{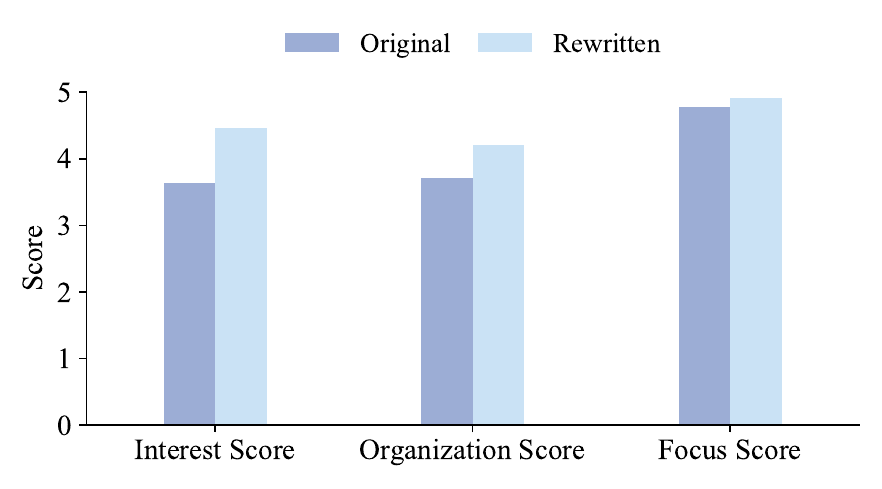}
    \caption{LLM-based evaluation results for human-written articles in the FreshWiki dataset.}
    \label{fig:human_written_comparison}
\end{figure}

\subsection{Configuration of the Memory Construction Module}

As described in Section~\ref{sec:memory_construction}, the memory construction stage iteratively refines a given topic into subtopics to extract relevant memory units. Several parameters influence the trade-off between time, cost, and the amount of curated information. These parameters include the maximum number of queries per topic, the maximum number of webpages processed, and the maximum depth of subtopic exploration. Based on preliminary empirical testing, we set the maximum queries to 2, the maximum webpages to 3, and the maximum subtopic depth to 2 to achieve an effective balance.

\section{Evaluation Details}
\label{appendix:evaluation}
\subsection{Entity and Numerical Data}
We evaluate the counts and recall of both named entities and numerical data against human-written articles. While spaCy’s Named Entity Recognition (NER) system categorizes certain numerical values—such as dates, times, and percentages—as entities, we isolate the numerical data to provide a more focused examination of these critical factual elements. Accurately capturing numerical details is essential for ensuring the precision and reliability of the generated text, making this separate evaluation a key part of our analysis.

\subsection{LLM-based metrics}
We employ Prometheus-2~\citep{kim2024prometheus}, a 7B-parameter evaluator LLM\footnote{\url{https://huggingface.co/prometheus-eval/prometheus-7b-v2.0}}, to assess the generated articles based on three metrics: Interest Score, Organization Score, and Focus Score. While \citet{shao-etal-2024-assisting} used Prometheus~\citep{kim2023prometheus}, a 13B-parameter model\footnote{\url{https://huggingface.co/prometheus-eval/prometheus-13b-v1.0}}, its shorter context window required trimming articles to 2,000 words. In contrast, the newer 7B version offers a longer context window that supports articles up to 10,000 words, thereby minimizing the need for such truncation. The detailed evaluation criteria are provided in Table~\ref{tab:evaluation_criteria}.

\begin{table*}[ht] \centering
    \resizebox{\textwidth}{!}{
        \Large
        \begin{tabular}{l||*{4}{c}||*{3}{c}||*{4}{c}}
        \toprule
        & \multicolumn{4}{c||}{\textbf{Informativeness}} & \multicolumn{3}{c||}{\textbf{Verifiability}} & \multicolumn{4}{c}{\textbf{Comparison with Human-written Articles}}  \\
        & \begin{tabular}[c]{@{}c@{}}Section \\ Count\end{tabular} & \begin{tabular}[c]{@{}c@{}}Word \\ Count\end{tabular} & \begin{tabular}[c]{@{}c@{}}Entity \\ Count\end{tabular} & \begin{tabular}[c]{@{}c@{}}Numerical \\ Count\end{tabular} & \begin{tabular}[c]{@{}c@{}}Citation \\ Recall\end{tabular} & \begin{tabular}[c]{@{}c@{}}Citation \\ Precision\end{tabular} & \begin{tabular}[c]{@{}c@{}}Citation \\ Rate\end{tabular} & ROUGE-1 & ROUGE-L & \begin{tabular}[c]{@{}c@{}}Entity \\ Recall\end{tabular} & \begin{tabular}[c]{@{}c@{}}Numerical \\ Recall\end{tabular}  \\
        \midrule
        MOG & \textbf{21.46} (7.78)& \textbf{2046.28} & \textbf{141.26} & \textbf{13.18} & 85.72 & 79.68 & \textbf{99.66} & \textbf{56.97} & \textbf{28.51} & \textbf{26.53} & \textbf{58.07} \\
        ~-w/o SE & 9.78 (5.32)& 1041.96 & 83.66 & 7.90 & 87.84 & 81.35 & 99.40 & 43.41 & 22.91 & 20.85 & 53.11 \\
        ~-w/o MO & 8.18 (8.18)& 1360.04 & 105.28 & 10.52 & \textbf{88.59} & \textbf{81.39} & 99.75 & 50.54 & 26.50 & 23.92 & 57.46 \\
        \bottomrule
    \end{tabular}
    }
    \caption{Full results of ablation study: "w/o SE" indicates the exclusion of the Subtopic Explorer, and "w/o MO" indicates the exclusion of the Memory Organization procedure.}
    \label{tab:full_ablation_studies}
\end{table*}

\subsection{Performance of Human-written article}
\label{appendix:human_written_article}
In Table~\ref{fig:llm_results}, we observe that human-written articles achieve low performance on LLM-based metrics. To fairly compare informativeness across systems, we retained all tabular data, which contributes significantly to factual content. However, these articles were scraped directly from Wikipedia without further processing, and their raw format—including markup, disjointed text, and structured elements like tables and lists—may have negatively affected scores related to fluency and organization. To test this, we lightly reformatted the articles using \texttt{gpt-4o-2024-08-06}, converting structured content into narrative text and removing empty or broken sections while keeping the rest intact. As shown in Figure~\ref{fig:human_written_comparison}, the Interest Score rose from 3.64 to 4.46 and the Organization Score from 3.71 to 4.13, while the Focus Score changed only slightly, from 4.78 to 4.91. This suggests that Interest and Organization scores are more sensitive to formatting quality, while Focus Score remains relatively robust. We note that this outcome reflects a general limitation of current LLM-based evaluation and is not specific to our setting. Importantly, the comparison across systems remains valid: both MOG and RAG generate articles under the same prompting framework and section-by-section generation strategy, ensuring consistent evaluation conditions across methods.

\section{Example of Generated Content}
\label{appendix:sample}

At the end of this appendix, we provide a sample article generated by the model on the topic of the "2023 SEA Games". This sample demonstrates the model’s ability to produce structured, coherent, and citation-rich content. The article opens with a lead section that outlines the main topic, followed by sections designated with `\#` for primary headings and subsections indicated by `\#\#` for more specific divisions. Additional subdivisions follow the same pattern. Citations are included to support the arguments presented and are formatted as numerical references in square brackets, such as [1] or [1,2].

\begin{figure*}[!t]
\begin{lstlisting}[caption={Prompts used in MOG}, label={lst:mog-prompts}, captionpos=b, linewidth=\textwidth]
class AtomicFactExtractor(dspy.Signature):
    """
    Extract atomic facts strictly about a specific topic from the given text.

    Guidelines:
    1. Be explicit with name entities and time; avoid using pronouns such as "he", "she", "it", "they", or "the...".
    2. If the text is not about the topic, return an empty list "[]".
    """

    topic = dspy.InputField(desc="Specific topic to extract facts about")
    text = dspy.InputField(desc="Text that may contain information about the topic")
    fact_list = dspy.OutputField(desc="List of atomic facts strictly about the topic")
    
class Outliner(dspy.Signature):
    """You are an experienced Wikipedian tasked with creating a general, one-level outline for a Wikipedia article section
    Guidelines:
    1. Provide only main subsection headings, without any further subdivisions.
    2. Ensure comprehensive coverage of key aspects related to the section title.
    3. Focus on the most important aspects within the section.
    4. Follow Wikipedia's style and naming conventions."""

    section_title = dspy.InputField(prefix="Section title: ", format=str)
    outline = dspy.OutputField(prefix="List of the subsection headings: ", format=str)


class OutlineRewriter(dspy.Signature):
    """You are an experienced Wikipedian tasked with improving an existing one-level outline for a Wikipedia article section.
    Guidelines:
    1. Provide only main subsection headings, without any further subdivisions.
    2. Maintain the original structure of the original outline.
    3. Add general headings, and delete unreasonable headings according to the collected information.
    4. Focus on the most important aspects within the section.
    5. Follow Wikipedia's style and naming conventions.
    """

    section_title = dspy.InputField(prefix="Section title: ", format=str)
    information_collected = dspy.InputField(
        prefix="The information collected", format=str
    )
    current_outline = dspy.InputField(prefix="Current outline:", format=list[str])
    outline = dspy.OutputField(prefix="List of the subsection headings: ", format=str)

class OutlineRefiner(dspy.Signature):
    """Refine the outline for the Wikipedia page.
    1. Remove redundant subsections.
    2. Keep all the sections.
    3. Do not add any new section."""

    outline = dspy.InputField(prefix="Outline: ")
    refined_outline = dspy.OutputField(prefix="Refined outline: ")
\end{lstlisting}
\end{figure*}

\begin{figure*}[!t]
\begin{lstlisting}[caption={Prompts used in MOG (continue)}, captionpos=b, label={lst:mog-prompts-continue}, linewidth=\textwidth]
class SectionWriter(dspy.Signature):
    """Write a Wikipedia section based on atomic facts. Do not improvise with any other information.
    1. Do not include section name as output.
    2. Ensure the section is split into multiple coherent paragraphs if necessary.
    3. The sequence of facts must be adjusted for coherence and readability.
    4. Stay focused within the section title, only include facts related to the section title.
    """

    topic = dspy.InputField(prefix="Topic: ", format=str)
    section_name = dspy.InputField(prefix="Section title: ", format=str)
    fact_list = dspy.InputField(prefix="Atomic facts: ", format=list[str])
    content = dspy.OutputField(prefix="Section content: ", format=str)

class SectionRefiner(dspy.Signature):
    """
    Refine the section of the given topic while keeping the markdown section headings unchanged.
    Stay focus on the section, remove unnecessary information about the section, and improve the coherence and writing.
    Only output the refined text, no other information.
    """

    topic = dspy.InputField(desc="Topic of the text")
    text = dspy.InputField(desc="Text to refine")
    refined_text = dspy.OutputField(desc="Refined text")
    
class CitationFinder(dspy.Signature):
    """Given a sentence and a source list, only cite the most relevant sources to fully cover the sentence.
    The output should be a list of indices of the sources in the input list.
    """

    claim = dspy.InputField()
    source_list = dspy.InputField(desc="a list of source text")
    answer = dspy.OutputField(
        type=list[int],
        desc="the indices of the most relevant source text, reply only the indices",
    )

class Entailer(dspy.Signature):
    """Is the claim faithful to the source? A claim is faithful to the source if the core part in the claim can be supported by the source.
    Start your answer with 'Yes' or 'No'."""

    source = dspy.InputField()
    claim = dspy.InputField()
    answer = dspy.OutputField(desc="reply only 'Yes' or 'No'")


class PartialEntailer(dspy.Signature):
    """Can the source at least partially support the claim?
    Start your answer with 'Yes' or 'No'.
    """

    source = dspy.InputField()
    claim = dspy.InputField()
    answer = dspy.OutputField(desc="reply only 'Yes' or 'No'")
\end{lstlisting}
\end{figure*}

\label{appendix:implementation}

\begin{table*}[ht]
\centering
\begin{tabularx}{1\textwidth}{|l|X|}
\hline
\textbf{Criteria} & \textbf{Description and Scoring Details} \\ \hline
\textbf{Interest Score} & 
\begin{itemize}[nosep]
    \item \textbf{Description:} How engaging and thought-provoking is the article?
    \item \textbf{Score 1:} Not engaging at all; no attempt to capture the reader's attention.
    \item \textbf{Score 2:} Fairly engaging with a basic narrative but lacking depth.
    \item \textbf{Score 3:} Moderately engaging with several interesting points.
    \item \textbf{Score 4:} Quite engaging with a well-structured narrative and noteworthy points that frequently capture and retain attention.
    \item \textbf{Score 5:} Exceptionally engaging throughout, with a compelling narrative that consistently stimulates interest.
\end{itemize} \\ \hline
\textbf{Organization Score} & 
\begin{itemize}[nosep]
    \item \textbf{Description:} Is the Wikipedia article structured according to Wikipedia's guidelines with clear sections and logical flow?
    \item \textbf{Score 1:} Poorly organized; lacks clear sections, headings are missing or inappropriate, and the structure does not follow Wikipedia guidelines.
    \item \textbf{Score 2:} Inadequately organized; some sections are present but lack clear headings or logical order, partially adhering to Wikipedia structure.
    \item \textbf{Score 3:} Adequately organized; includes standard sections with mostly appropriate headings and a logical flow, follows Wikipedia structure with minor issues.
    \item \textbf{Score 4:} Well organized; clear and appropriate sections and headings, logical flow, adheres to Wikipedia organizational standards with few minor lapses.
    \item \textbf{Score 5:} Excellently organized; perfectly structured according to Wikipedia guidelines with clear, appropriate sections and headings, seamless logical flow.
\end{itemize} \\ \hline
\textbf{Focus Score} & 
\begin{itemize}[nosep]
    \item \textbf{Description:} Does the article stay on topic and maintain a clear focus?
    \item \textbf{Score 1:} Off-topic; the content does not align with the headline or core subject.
    \item \textbf{Score 2:} Somewhat on topic but with several digressions; the core subject is evident but not consistently adhered to.
    \item \textbf{Score 3:} Generally on topic, despite a few unrelated details.
    \item \textbf{Score 4:} Mostly on topic and focused; the narrative has a consistent relevance to the core subject with infrequent digressions.
    \item \textbf{Score 5:} Exceptionally focused and entirely on topic; the article is tightly centered on the subject, with every piece of information contributing to a comprehensive understanding of the topic.
\end{itemize} \\ \hline
\end{tabularx}
\caption{Evaluation Criteria for Generated Articles}
\label{tab:evaluation_criteria}
\end{table*}

\clearpage
\onecolumn
\noindent\textbf{2023 SEA Games} \\\\
The 2023 Southeast Asian Games, commonly known as the SEA Games, were held from May 5 to May 17, 2023.[6,8,9,10,11] This edition marked the first time Cambodia hosted the event, with Phnom Penh serving as the main hub.[1,6,10,13,14,15] The games also took place in other cities across Cambodia, including Siem Reap, Sihanoukville, Kep, and Kampot provinces.[1] The Morodok Techo National Stadium in Phnom Penh was a focal point, hosting the opening and closing ceremonies.[1] \\\\
\noindent The 2023 SEA Games featured over 7,000 athletes from 11 Southeast Asian countries, competing in approximately 40 sports and a record 632 events, making it the largest in SEA Games history.[1,3,4,5,6] Notable achievements included Team Singapore winning a total of 51 gold medals, with Quah Ting Wen securing six golds.[9,12] Cambodia's team also excelled, winning 81 gold medals, a significant increase from their previous best.[7] \\\\
\noindent The event was not only a sporting spectacle but also a significant economic opportunity for Cambodia.[5,16,17] Hosting the games was expected to generate up to \$400 million in economic activity, create new jobs, and improve the country's GDP.[5,16] All events were free for both locals and foreigners, fostering a sense of national unity and pride among Cambodians.[2,4,9] The theme of the games was "Sports Live in Peace," reflecting the spirit of camaraderie and competition.[2] \newline

\noindent\textbf{\# History of the SEA Games}\\
\noindent\textbf{\#\# Origins and Inception}\\
The Southeast Asian Games were conceived by Luang Sukhum Nayapradit in 1958 during the Third Asian Games in Tokyo.[1,2,3] He envisioned a smaller sports event for Southeast Asian nations, leading to a meeting on May 22, 1958, with representatives from Burma (now Myanmar), Laos, and Malaysia in Thailand.[1] The first Southeast Asian Peninsular Games, later known as the Southeast Asian Games, took place in Bangkok, Thailand, in 1959, featuring around 800 athletes and officials in 12 sports.[1,4] Participating countries included Thailand, Burma, Malaya (now Malaysia), Singapore, South Vietnam, and Laos.[1,3] General Prabhas Charusatiara of Thailand was the first president of the Southeast Asian Peninsula Games Federation, established in June 1959 with founding members Burma, Cambodia, Laos, Malaysia, Thailand, and Vietnam.[1] In 1969, Singapore proposed renaming the SEAP Games to the SEA Games to include more countries like Indonesia and the Philippines.[1,3] This change was officially adopted in 1977 with the inclusion of Brunei, the Philippines, and Indonesia.[3] The Games have since expanded to feature over 40 sports and more than 10,000 athletes.[4] \newline

\noindent\textbf{\#\# Renaming and Broadened Significance}\\
The 2023 SEA Games highlighted the renaming and cultural significance of certain sports.[2,5,6,7,8] The change from Muay Thai to Kun Khmer for the kickboxing event sparked a cultural debate between Cambodia and Thailand.[2,6,8,9] Cambodia introduced 'Kun Khmer', also known as Pradal Serey, replacing 'Muay Thai'.[2,6,10] Despite criticism, Muay Thai was recognized by the International Olympic Committee, and at the SEA Games, it was referred to by its Khmer name, following Cambodian rules.[2,8] The renaming reflects historical context and Cambodia's cultural heritage.[5,6,7] In 1995, Cambodia suggested 'Sovannaphum boxing' or 'SEA Boxing', with 'Sovannaphum' meaning 'golden land' in Khmer.[5] Kun Khmer represents Cambodia's national pride.[5,6,7] The 2023 SEA Games symbolized healing and unity for Cambodia, enhancing its regional reputation.[7,11] As the event concluded, the SEA Games federation flag was passed to Thailand, the next host in 2025.[12] \newline

\noindent\textbf{\#\# Major Events and Competitions}\\
The 2023 Southeast Asian Games showcased diverse events, including nine esports competitions, highlighting the rise of competitive gaming.[13,14] The Mobile Legends: Bang Bang Men's Tournament was a notable event, marking its third appearance as a medal event in a multi-sport competition.[14] Athletics featured memorable performances, such as Indonesia's Hendro Yap winning gold in the men's 20-km fast-walk event, despite weather-related challenges.[2] In the women's 5k, Vietnam's Thi Oanh won, while Cambodia's Bou Samnang inspired many by finishing despite adverse conditions.[2] Traditional sports included the women's marathon, where Indonesia's Odekta Elvina Naibaho triumphed.[15] The Philippines' aquathlon team won gold in the mixed relay, underscoring the athletes' competitive spirit and dedication.[16] \newline

\noindent\textbf{\#\# Role in Regional Integration and Unity}\\
The SEA Games have significantly contributed to regional integration and unity among ASEAN nations.[6] All ASEAN members, except Timor Leste, have hosted the Games, emphasizing its role in promoting cooperation and camaraderie.[17] The Southeast Asian Games Federation regulates the event, ensuring alignment with the shared values of participating countries.[6] The Games share characteristics with ASEAN, focusing on informality and consensus, key aspects of ASEAN's regional diplomacy.[6] Supervised by the International Olympic Committee and the Olympic Council of Asia, the Games adhere to international standards, maintaining competition integrity and enhancing regional unity through sports.[6,18] \newline

\noindent\textbf{\# Host City and Venues}\\
\noindent\textbf{\#\# Host City Selection}\\
The 2023 Southeast Asian Games were hosted by Phnom Penh, Cambodia, marking the first time Cambodia hosted the event.[6,14,17,19,20,21] Initially, the Philippines was set to host, but the event was rescheduled for Cambodia following Brunei’s withdrawal.[22,23,24] The SEA Games Federation Council awarded Cambodia the hosting rights in 2015.[14,17,21,24,25] The games took place from May 5-17, 2023.[21] Cambodia invested \$30–40 million in facilities and provided free lodging for over 12,000 athletes and delegations.[7] This event was a significant milestone for Cambodia, with 53.1\% of Cambodians expressing interest in attending.[6,14,17,21] Cambodia will not host the SEA Games again until at least 2035.[26] \newline

\noindent\textbf{\#\# Major Venues and Facilities}\\
The Morodok Techo National Stadium in Phnom Penh was the centerpiece of the 2023 SEA Games.[7,27,28] Designed to resemble a sailing ship, it has a seating capacity of 60,000 and features sustainable designs like a green roof and rainwater systems.[24,27,29] Construction began in 2017 and was completed in 2021, funded by China at a cost of approximately \$150 million.[27,29,30,31] The Morodok Techo National Sports Complex spans 16 hectares and includes an aquatic center, a multi-sports gymnasium, and various indoor sports halls.[30,32,33,34] The complex also features the WorldBridge Sport Village, providing 6,100 beds for athletes and officials, and an Athletes' Village with a capacity of 7,000 residents.[27,30,33,35] \newline

\noindent\textbf{\#\# Satellite Venues: Siem Reap, Kep, Kampot, Sihanoukville}\\
In addition to Phnom Penh, the 2023 SEA Games extended to Siem Reap, Preah Sihanouk, Kampot, and Kep.[21,36,37] These locations were chosen to host various events, highlighting Cambodia's diverse regions.[21,36,37] Siem Reap, near the Angkor Wat temple complex, and the coastal provinces of Preah Sihanouk, Kampot, and Kep provided unique settings and facilities, enhancing the overall experience of the games.[21,36,37]\newline

\noindent\textbf{\# Participants and Teams}\\
\noindent\textbf{\#\# Countries Participating}\\
The 2023 Southeast Asian Games featured participation from 11 nations, all members of the Southeast Asian Games Federation: Brunei, Cambodia, Indonesia, Laos, Malaysia, Myanmar, the Philippines, Singapore, Thailand, Timor-Leste, and Vietnam.[7,26,38,39] This consistent lineup underscores the regional cooperation and sporting spirit the SEA Games aim to foster.[3,7,38,39] Over 6,000 athletes from these countries competed, with Cambodia as the host nation.[7,16] The inclusion of Kun Khmer, with participation from seven ASEAN countries, highlights the diverse sporting culture celebrated within the SEA Games framework.[2,9] \newline

\noindent\textbf{\#\# Participating Sports}\\
The Games, held in Cambodia, included over 7,000 athletes competing in 37 events, with a total of 40 sports.[25,40] Athletes vied for 5,151 medals, showcasing the region's athletic talent and fostering camaraderie among nations.[7,40] \newline

\noindent\textbf{\#\# Team Sizes and Composition}\\
The Men's Tournament featured nine teams divided into two groups.[14] Group A had five teams, while Group B had four, following a Single Round Robin format.[14] The tournament then progressed to single elimination playoffs.[14] Teams could enter up to seven events, but Cambodia, as the host, could enter all nine.[13] The Games included a minimum of 22 sports, ensuring a comprehensive event for all nations.[6] \newline

\noindent\textbf{\#\# Notable Teams and Athletes}\\
The Games showcased talented athletes and teams, including Neeraj Chopra, an Olympic and world champion in athletics, and Nikhat Zareen in boxing.[41] Other notable athletes were Izaac Quek Yong in table tennis, Jonathan Tan Eu Jin in swimming, and Loh Kean Yew in badminton.[38] The Gilas Pilipinas basketball team from the Philippines featured players like CJ Perez and Justin Brownlee.[39] Cambodia prepared by sending athletes for intensive training, although they imposed restrictions on athlete numbers in certain sports.[6,25] Overall, over 6,000 athletes from 11 nations competed in 584 events across 36 sports, highlighting the scale and competitive spirit of the Games.[16] \newline

\noindent\textbf{\# Sports and Events}\\
\noindent\textbf{\#\# Athletic Events}\\
The 2023 Southeast Asian (SEA) Games in Phnom Penh featured 36 sports with 584 events, highlighting regional athletic talent.[16] A standout moment was the men's 4x400m relay team winning gold.[39] Events were held across various locations, with Phnom Penh as the central hub.[32,42] The marathon events, starting and finishing at Angkor Wat, were particularly notable.[17] Traditional sports like Kun Khmer added cultural depth, while modern sports, including nine esports events like VALORANT, showcased the Games' evolving nature and dynamic spirit.[2,5,13] \newline

\noindent\textbf{\#\# Aquatic Sports}\\
Aquatic sports at the 2023 SEA Games highlighted regional talent in swimming and related disciplines.[3,4,6] Nguyen Thuy Hien earned a bronze in the women's 100m freestyle, celebrated alongside her idol, Quah Ting Wen.[43] The aquathlon mixed relay was a highlight, with a gold medal win underscoring the athletes' strength and teamwork.[39] Swimming remains a strong sport in Southeast Asia, with Singapore consistently producing top-tier swimmers.[4,38] The Games continued this tradition, enhancing the region's aquatic sports reputation.[3,4] \newline

\noindent\textbf{\#\# Team Sports}\\
The 2023 SEA Games in Cambodia featured diverse team sports, including football, basketball, and volleyball.[17,24,37] The men's football tournament showcased competitive spirit with nine teams.[14] Esports, now an official discipline, included games like Crossfire, League of Legends: Wild Rift, Mobile Legends: Bang Bang, PUBG Mobile, and Valorant.[42] Other team sports included hockey, floorball, and traditional boat racing, promoting cultural exchange and regional sports.[24,36] The event celebrated Southeast Asia's rich sporting culture and camaraderie.[18,24] \newline

\noindent\textbf{\#\# Paralympic Events}\\
The 2023 SEA Games marked the third time esports was a medal event, reflecting its growing importance.[14,19] The Games featured men's football and cricket competitions, with notable matches in the Men's and Women's Twenty20 Cricket Competitions.[44,45,46] In esports, Singapore and Indonesia competed in the VALORANT finals, highlighting esports' recognition within the SEA Games.[2,14,19] The inclusion of esports as a medal event underscores the evolving sports landscape and the rising popularity of competitive gaming in the region.[14,19] \newline

\noindent\textbf{\# Opening Ceremony}\\
\noindent\textbf{\#\# Cultural Performances}\\
The opening ceremony of the 2023 Southeast Asian Games took place on May 5, 2023, at the Morodok Techo National Stadium in Phnom Penh, Cambodia.[2,28,42,47,48] This grand event marked the official start of the 32nd SEA Games, attracting tens of thousands of spectators.[28,47,48] Designed to be of Olympic standard, the ceremony featured cultural performances that highlighted Cambodia's rich history and vibrant culture.[7,37,49] The event was divided into three parts, each showcasing sound and lighting performances.[37,49] A significant highlight was a 45-minute extravaganza entitled "Cambodian Journey," which included choreographed performances and colorful floats celebrating Cambodia's history and culture.[49] The parade of athletes was a key feature, with representatives from 11 Southeast Asian countries participating.[2,28,48] Notably, Team Philippines wore Francis Libiran's "Araw" barong ensemble during the parade, and Alyssa Valdez served as the Philippines’ flag bearer.[28,47] The ceremony concluded with the lighting of the Games cauldron by Cambodia's Sorn Seavmey, followed by a two-minute fireworks display.[47,49] The official anthem of the SEA Games 32, "Cambodian Pride," was played to close the ceremony.[37] The message of the opening ceremony, "Sports, live in peace," resonated throughout the event, setting a tone of unity and celebration for the games ahead.[37,47,48] \newline

\noindent\textbf{\#\# Key Highlights}\\
The opening ceremony of the 2023 SEA Games was held on May 5, marking the beginning of the regional multi-sport competition and bringing together athletes from across Southeast Asia.[2] The ceremony was a vibrant display of culture and unity, setting the stage for the games that followed.[2] A notable highlight was the parade of athletes, featuring prominent sports figures such as Kristina Knott, Rubilen Amit, Afril Bernardino, Sarina Bolden, and Rianne Malixi.[28] These athletes joined Alyssa Valdez in representing their countries, showcasing the diverse talent present at the games.[28] The countdown to the 2023 SEA Games began much earlier, with a ceremony presided over by Prime Minister Hun Sen on December 18, 2021, signifying the anticipation and preparation leading up to the games.[25] \newline

\noindent\textbf{\# Closing Ceremony}\\
The closing ceremony of the 2023 SEA Games was held on May 17, 2023, at the Morodok Techo National Stadium in Phnom Penh, Cambodia.[18,50] Presided over by Cambodia's Prime Minister, Samdech Akka Moha Sena Padei Techo Hun Sen, the event marked the official conclusion of the Games.[50,51,52] The 1.5-hour ceremony featured vibrant lighting, song and dance performances, and a martial arts segment showcasing bokator, a traditional Cambodian martial art.[50] Despite a thunderstorm, performers continued to sing, dance, and cheer, enhancing the lively atmosphere.[50] The SEA Games flame was extinguished, and the SEAGF Flag Lowering Ceremony took place.[12,18,51] Awards were distributed, including the Best Athlete award to Quah Ting Wen from Singapore.[12,51,53] The ceremony concluded with a musical performance titled 'Cambodian Pride' and a four-minute fireworks display.[50] The SEA Games flag was handed over to Thailand, the next host of the Games.[50] The event was broadcast live, allowing audiences across the region to witness the culmination of the 32nd SEA Games.[54] \newline

\noindent\textbf{\# Medal Tally}\\
The 2023 Southeast Asian Games saw a record-breaking 5,151 medals awarded across 37 events, with 608 sets of medals available, the highest in the history of the Games.[36,40] Host nation Cambodia excelled, winning 282 medals, including 81 gold, 74 silver, and 127 bronze.[51] The Philippines also performed admirably, securing 261 medals, with 58 gold, 87 silver, and 116 bronze, placing them fifth overall.[55] Singapore's athletes contributed 44 medals to their tally, highlighted by 13 silver medals.[53] Esports emerged as a significant category, initially offering 27 medals, but the tally increased to 33 due to shared medals, underscoring its growing importance.[13] Notable individual achievements included Singapore's Quah Ting Wen, who won six gold medals, and the Philippines' Trixie Lofranca, who earned a gold in arnis, showcasing the diverse talents at the Games.[39,53] \newline

\noindent\textbf{\# Notable Achievements and Records}\\
\noindent\textbf{\#\# Overview of Records Set}\\
During the 32nd SEA Games, athletes set seventeen national records and eight Games records, demonstrating exceptional performances and competitive spirit.[12,56] Additionally, forty personal bests were achieved, highlighting individual accomplishments.[56] In cricket, standout performances included Thailand's top female run scorer with 115 runs in 4 innings, Malaysia's top scorer with 158 runs in 3 innings, and Singapore's top scorer with 159 runs in 3 innings.[45,46] Cambodia's top scorer also impressed with 117 runs in 3 innings.[45] Among the top wicket-takers, Thailand Women's Sla took 4 wickets in 11 innings, while Malaysia's Rmf and Ob players took 3 and 4 wickets, respectively, showcasing the high level of competition.[46] \newline

\noindent\textbf{\#\# Team Performance Highlights}\\
The 2023 SEA Games featured remarkable achievements across various sports.[39,53,56,57,58] In gymnastics, Juancho Miguel Besana from the Philippines scored 73.700 in the all-around, while Suphacheep Baobenmad from Thailand scored 66.050.[57] In swimming, Singapore's Quah Zheng Wen won gold in the men's 100m backstroke and butterfly, and Laetitia Sim secured gold in the women's breaststroke and medley events.[58] Quah Ting Wen won her fifth consecutive gold in the women's 100m freestyle, and Jonathan Tan achieved gold in the men's 50m and 100m freestyle, meeting the Olympic 'A' cut.[53,56,58] Singapore's relay teams excelled, winning gold in multiple events.[53,56,58] Other notable achievements included Aloysius Yapp's silver in men's 9-ball pool and Ang Chen Xiang's gold in the men's 110m hurdles.[58] Singapore concluded with the best overall medal tally in swimming, with 47 medals, including 22 golds.[56] \newline

\noindent\textbf{\#\# Noteworthy National Achievements}\\
The Games saw significant achievements in esports, with the Philippines' Sibol team winning gold in League of Legends: Wild Rift, and Singapore and Indonesia sharing gold in VALORANT.[2,13,16,39] Indonesia also excelled in PUBG Mobile and MLBB Women's categories, setting a viewership record.[13] In athletics, Singapore's Shanti Pereira won gold in the women's 100m and 200m, while Puripol Boonson secured gold in the men's sprints.[58,59] The Philippines' EJ Obiena captured gold in men's pole vault.[16] In traditional martial arts, Indonesia and Vietnam shared gold in women's pencak silat, and the Philippines excelled in arnis.[2] In gymnastics, Carlos Edriel Yulo from the Philippines scored 84.000 in the all-around, with Vietnam's Le Thanh Tung ranking second.[57] The Games concluded with Vietnam leading the medal standings, followed by Thailand and Indonesia, while host nation Cambodia finished fourth.[50,57] \newline

\noindent\textbf{\# Controversies and Challenges}\\
\noindent\textbf{\#\# Sports Fairness and Regulations}\\
The 2023 Southeast Asian (SEA) Games encountered several controversies, particularly regarding sports fairness and regulations.[6,36] The SEA Games, overseen by the Southeast Asian Games Federation and supervised by the International Olympic Committee (IOC) and the Olympic Council of Asia, require a minimum of 22 sports.[6,27] However, host countries can modify the sports lineup, leading to disputes.[6,36] In 2023, Cambodia, the host nation, excluded several Olympic sports like shooting and archery, sparking debates about fairness and inclusivity.[6,36] Concerns arose about Cambodia's coordination with international governing bodies, essential for hosting international events.[8,27] Muay Thai, recognized by the IOC, must be supervised by the International Federation of Muaythai Associations (IFMA) for official tournaments.[8] This underscores the need for adherence to international standards to ensure fairness.[8,27] The SEA Games' preference for informality can challenge consistent regulations, with past controversies including wrongful judging and match-fixing, highlighting the need for stringent oversight.[6] \newline

\noindent\textbf{\#\# Infrastructure and Logistical Issues}\\
The 2023 SEA Games faced significant infrastructure and logistical challenges, affecting the event's execution.[2] Faulty infrastructure led to issues like track events finishing in darkness, raising concerns about facility adequacy.[2] Athletes from the Philippines and Singapore reported injuries from subpar basketball courts.[2] Despite projected hosting expenses exceeding \$150 million, infrastructure issues persisted, drawing criticism.[2,7] The Morodok Techo National Stadium passed inspections, indicating varied scrutiny levels across venues.[2,32] These challenges echo past SEA Games issues, emphasizing the need for improved planning and investment to ensure athlete and spectator safety and satisfaction.[2] \newline

\noindent\textbf{\#\# Cultural Controversies and Political Tensions}\\
The 2023 SEA Games were marked by cultural controversies and political tensions, notably between Thailand and Cambodia.[6,10,60] A major issue was Cambodia's renaming of the kickboxing event from 'Muay Thai' to 'Kun Khmer,' opposed by Thailand, which threatened a boycott.[6,10] The International Federation of Muaythai Associations (IFMA) criticized this decision.[5,6,9] Tensions escalated when Thailand rejected a broadcasting rights deal over increased fees.[2,6,61] Criticism also arose over Cambodia's regulatory decisions perceived as attempts to boost its medal tally, with restrictions on athlete participation in certain sports not applied to itself.[6] The opening ceremony faced backlash for displaying flags of Indonesia, Vietnam, and Myanmar upside down, prompting an apology.[2,50] Additionally, the esports competition saw Indonesia accuse Singapore of 'digital doping' during the VALORANT finals, leading to Indonesia's forfeiture.[2,43] These controversies highlighted the challenges in maintaining fairness and goodwill, with tensions between Thailand and Cambodia underscoring the complex interplay of cultural pride and political interests in international sports.[6,10,60]\newline

\noindent\textbf{\# Sponsorship and Broadcasting}\\
\noindent\textbf{\#\# Sponsorship Overview}\\
Ajinomoto was an official sponsor of the 32nd Southeast Asian Games, continuing its support since the 29th SEA Games in 2017.[62] TikTok, as a premium sponsor, promoted SEA Games content and supported event broadcasting, enhancing visibility through its platform.[61,62] Cambodia Airways provided crucial transport services for delegations, ensuring smooth logistics for athletes and officials.[61] \newline

\noindent\textbf{\#\# Broadcasting Rights and Coverage}\\
Cambodian Sports Television (CSTV) broadcasted the 2023 SEA Games domestically, in collaboration with the General Department of National Television of Cambodia.[61,63] The Southeast Asian Games Federation granted Cambodia the broadcasting rights, with CSTV offering free live streams on its YouTube channel.[26,63] The National Olympic Committee of Cambodia established an International Broadcasting Centre to facilitate coverage.[31] Broadcasting rights were acquired in Vietnam, Singapore, Indonesia, Malaysia, and Cambodia.[61,63,64] VTVcab in Vietnam, Mediacorp in Singapore, and MNC Media in Malaysia secured the rights, with Mediacorp focusing on Team Singapore.[40,54,61,63,65] Cambodia set the broadcast rights price at \$800,000 for Thailand, advocating a profit-sharing model.[6,25] Apollo Media and On Media distributed rights in Vietnam, with 50\% of regional rights sold by March 2023.[54,61] Mediacorp used advanced technology for remote live production, enhancing international commentary.[40] The media presence included over 2,070 individuals, highlighting significant interest.[35] \newline

\noindent\textbf{\#\# Advertising and Marketing Campaigns}\\
The 2023 SEA Games featured diverse advertising and marketing campaigns.[62] Brands engaged in experiential marketing at venues, creating interactive experiences to connect with attendees.[62] Digital campaigns were prominent, with Milo and MNC Vision launching Instagram initiatives offering prizes.[62] Endorsements and special edition merchandise were key strategies, with athletes like Fajar Faturahman receiving endorsements, and Jakarta Warriors unveiling special edition jerseys celebrating the event.[62] \newline

\noindent\textbf{\# Cultural and Economic Impact}\\
\noindent\textbf{\#\# Cultural Significance and Promotion of National Identity}\\
The Southeast Asian Games (SEA Games) foster unity and cooperation among Southeast Asian countries, promoting cultural exchange and national pride.[26] The 2023 SEA Games in Cambodia, with the slogan 'Sport: Live In Peace,' emphasized peace and harmony through sports.[25,26] The event serves as a platform for showcasing national identity, allowing host nations to highlight their cultural heritage.[26] Traditional sports like Muay Thai enhance cultural visibility and contribute to national pride.[9,10] \newline

\noindent\textbf{\#\# Economic Impact and Investment Opportunities}\\
The SEA Games significantly impact host countries' economies by creating investment opportunities and boosting local economies.[11,21,24,66] Hosting the 2023 SEA Games is expected to generate substantial economic activity for Cambodia, driven by increased demand for local goods and services.[24] Cambodia invested approximately USD 30 million in preparation, with total hosting costs estimated between \$100 million and \$200 million.[11,21] These investments are crucial for infrastructure development and accommodating visitors.[21] The experience of the Philippines during the 2019 SEA Games demonstrated the potential economic benefits, with significant contributions from regions like Clark City.[66,67] Government support, such as the additional P1 billion approved by then-President Rodrigo Duterte, underscores the importance of maximizing economic benefits.[67] \newline

\noindent\textbf{\#\# Tourism Development and International Visitors}\\
The 2023 SEA Games are expected to boost tourism in Cambodia, promoting it as a tourist destination.[11,24] The Cambodian Tourism Minister projected about 500,000 foreign arrivals during the Games.[11] Free admission to events aims to encourage attendance and participation, enhancing cultural exchange and economic benefits.[42] The availability of online ticket purchases facilitates access for international tourists.[31] The impact on tourism extends beyond Cambodia, as seen in the 2019 SEA Games in the Philippines, which positively affected local tourism industries.[66] The diverse visitor demographics reflect regional interest and potential economic partnerships, stimulating local businesses and contributing to Cambodia's economic development.[11] \newline

\noindent\textbf{\# Legacy and Future Implications}\\
\noindent\textbf{\#\# Infrastructure and Economic Impact}\\
The 2023 Southeast Asian Games have left a significant legacy in Cambodia, particularly in infrastructure and economic impact.[7,11,21,31] Hosting the games required substantial investment, estimated between \$100 million and \$200 million, including \$30–40 million for constructing and equipping facilities.[7,11,21] This investment generated demand for local goods and services, created jobs, and improved Cambodia's GDP, bringing both prestige and economic benefits.[11,24,31] The games also spurred long-term infrastructure projects, such as a new international airport in Siem Reap and an expressway connecting Phnom Penh and Siem Reap, enhancing transportation and logistics.[11] These developments are part of a broader effort to modernize the country's infrastructure, with the expressway expected to improve connectivity between major cities.[11] The legacy extends beyond infrastructure, positively impacting Cambodia's property market and increasing the rights fee to almost \$900,000, reflecting the event's growing significance.[21,31,61] Overall, the investments and developments associated with the 2023 SEA Games are set to benefit future generations in Cambodia.[7,11,21,31] \newline

\noindent\textbf{\#\# Athletic Achievements and Future Prospects}\\
The 2023 Southeast Asian Games, hosted by Cambodia for the first time, marked a significant milestone in the region's sporting history.[25,26,47] Phnom Penh served as the host city, with events also in Siem Reap.[19,26,42] The Games featured approximately 8,000 athletes competing across 40 sports.[25] The Cambodian government invested around \$400 million to organize the event and construct the Morodok Techo National Stadium.[31] Notable athletic achievements included Eric Cray's 400-metre hurdles title and Sibol's gold medal in esports.[16,55] Indonesia's football team triumphed over Thailand in the final match.[2] Cambodia's national football team reached the semifinal bronze medal match but lost to Myanmar.[25] Despite challenges, football remained the most popular sport among Cambodians.[21] The focus on international competition has been strengthened to improve Cambodia's chances in future events.[25] The 12th ASEAN Para Games were also held in Cambodia, highlighting the country's growing role in regional sports.[25,42] Attention now shifts to Thailand, the next host for the 2025 SEA Games.[3,68] \newline

\noindent\textbf{\#\# Social Impact and Community Engagement}\\
The 2023 SEA Games had a significant social impact and fostered community engagement across Southeast Asia.[20,52,62] The event promoted sports and encouraged young people to adopt an active lifestyle.[20] Initiatives like Milo's digital campaign and the Ajinomoto Group's ASEAN Victory Project supported athletes and highlighted community backing in sports.[62] Cambodia showcased its cultural heritage, particularly Kun Khmer, a traditional martial art recognized by UNESCO.[9,10,52] Despite Muay Thai's larger following, efforts to promote Kun Khmer are increasing, contributing to Cambodia's cultural renaissance.[10] The Morodok Techo National Stadium symbolized the strengthening friendship between Cambodia and China, part of China's Belt and Road initiative.[32] This collaboration is expected to have long-term implications.[10,31,32] The Games also highlighted the flexibility of host countries in shaping the event, allowing for unique cultural expression.[6,9] Despite challenges, such as the mishandling of flags during the opening ceremony, the Games successfully fostered unity and cultural exchange across the region.[2,9,20]
\twocolumn

\end{document}